\documentclass[journal]{IEEEtran}

\ifCLASSINFOpdf
    \usepackage{graphicx}
    \usepackage{epstopdf}
    \usepackage{array}
    \usepackage{multirow}
    \usepackage{float}
    \usepackage{subfigure}
    \usepackage{times}
    \usepackage{bm}
    \usepackage{tabularx}
    \usepackage{color}
    \usepackage{amsfonts}
    \usepackage{hyperref}
    \usepackage[justification=centering]{caption}
    \newcolumntype{I}{!{\vrule width 0.7pt}}
    
    \newlength\savewidth
    \newcommand\shline{\noalign{\global\savewidth\arrayrulewidth
                           \global\arrayrulewidth 0.7pt}%
                  \hline
                  \noalign{\global\arrayrulewidth\savewidth}}
\else
\fi

\begin{document}

\title{Fine-Grained 3D Shape Classification with Hierarchical Part-View Attention}

\author{Xinhai Liu,
        Zhizhong Han,
        Yu-Shen Liu,
        and Matthias Zwicker
\thanks{X. Liu and Y.-S. Liu are with the School of Software, BNRist, Tsinghua University, Beijing 100084, China (e-mail: lxh17@mails.tsinghua.edu.cn, liuyushen@tsinghua.edu.cn)(Corresponding author: Yu-Shen Liu).}
\thanks{Z. Han and Matthias Zwicker are with the University of Maryland, College Park, 20737, USA (e-mail: h312h@mail.nwpu.edu.cn, email: zwicker@cs.umd.edu).}
\thanks{This work was supported by National Key R\&D Program of China (2020YFF0304100, 2018YFB0505400), the National Natural Science Foundation of China (62072268), in part by Tsinghua-Kuaishou Institute of Future Media Data, and NSF (award 1813583).}
}

\maketitle

\begin{abstract}
Fine-grained 3D shape classification is important for shape understanding and analysis, which poses a challenging research problem. 
However, the studies on the fine-grained 3D shape classification have rarely been explored, due to the lack of fine-grained 3D shape
benchmarks.
To address this issue, we first introduce a new 3D shape dataset (named FG3D dataset) with fine-grained class labels, which consists of three categories including airplane, car and chair. 
Each category consists of  several subcategories at a fine-grained level.
According to our experiments under this fine-grained dataset, we find that state-of-the-art methods are significantly limited by the small variance among subcategories in the same category.
To resolve this problem, we further propose a novel fine-grained 3D shape classification method named FG3D-Net to capture the fine-grained local details of 3D shapes from multiple rendered views.
Specifically, we first train a Region Proposal Network (RPN) to detect the generally semantic parts inside multiple views under the benchmark of generally semantic part detection.
Then, we design a \textit{hierarchical part-view attention aggregation} module to learn a global shape representation by aggregating generally semantic part features, which preserves the local details of 3D shapes.
The part-view attention module hierarchically leverages part-level and view-level attention to increase the discriminability of our features. 
The part-level attention highlights the important parts in each view while the view-level attention highlights the discriminative views among all the views of the same object.
In addition, we integrate a Recurrent Neural Network (RNN) to capture the spatial relationships among sequential views from different viewpoints.
Our results under the fine-grained 3D shape dataset show that our method outperforms other state-of-the-art methods.
The FG3D dataset is available at \url{https://github.com/liuxinhai/FG3D-Net}.
\end{abstract}


%
\IEEEpeerreviewmaketitle

\section{Introduction}
\IEEEPARstart{L}{earning} a shape representation from multiple rendered views is an effective way to understand 3D shapes \cite{Su_2015_ICCV,bai2017gift,Kanezaki_2018_CVPR,han2019y2seq2seq}.
Influenced by the great success of Convolutional Neural Networks (CNNs) in the recognition of 2D images under large-scale datasets, such as ImageNet \cite{deng2009imagenet}, 2D CNNs are intuitively applied to learn the representation for 3D shapes.
For example, the pioneering MVCNN \cite{Su_2015_ICCV} first projects a 3D shape into multiple views from different viewpoints and then obtains a global 3D shape representation by aggregating the view features with a view pooling layer, where the view features are extracted by a shared CNN.
Previous view-based methods have achieved satisfactory performance for 3D shape recognition under large variance among different categories.
However, it is still nontrivial for these methods to capture the small variance among subcategories in the same category, which limits the discriminability of learned features for fine-grained 3D shape recognition.
\begin{figure}
    \centering
    \includegraphics[height=5cm]{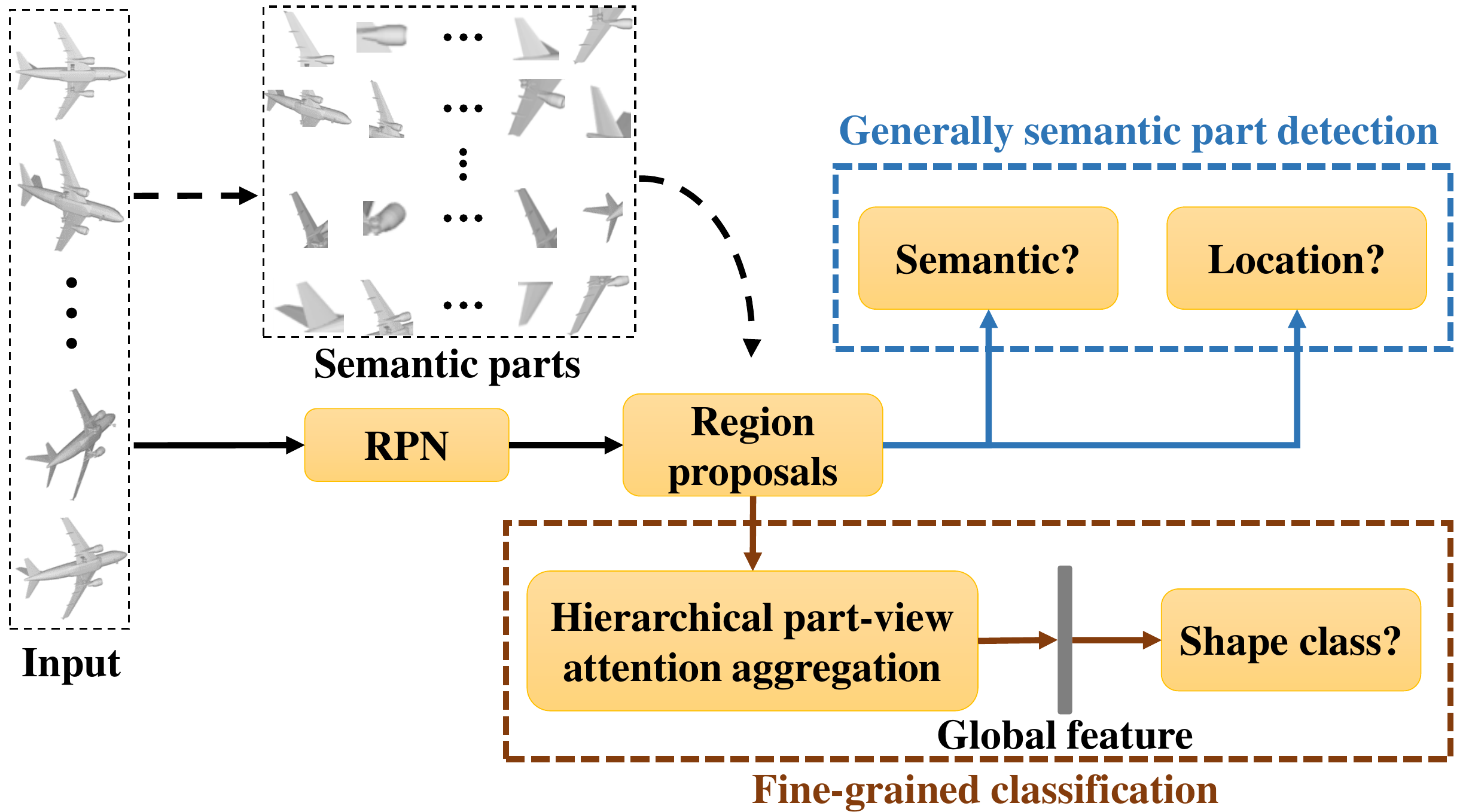}
    \caption{The illustration of our FG3D-Net.
    In FG3D-Net, we first render each 3D shape into multiple views that are propagated into a Region Proposal Network (RPN) \cite{ren2015faster} to generate a set of region proposals.
    Then, the extracted region proposals are used in two different branches including generally semantic part detection (top branch in blue box) and fine-grained classification (bottom branch in brown).
    In the generally semantic part detection branch, we predict the semantic score and the bounding box location for each region proposal.
    According to the semantic scores, several region proposals are selected to extract the global feature of the input 3D shape in the fine-grained classification branch.
    Specifically, we introduce a novel module named \textit{hierarchical part-view attention aggregation} to effectively capture the fine-grained 3D shape details for fine-grained classification.
    }
    \label{fig:idea}
\end{figure}

Fine-grained 3D shape recognition, which aims to discriminate 3D shapes within the same category, such as airliners, fighters and seaplanes within the airplane category, is quite challenging.
Specifically, there are two major issues that limit the performance of fine-grained classification of 3D shapes.
On the one hand, many large-scale image datasets (e.g. CUB-200-2011 \cite{wah2011caltech} and Stanford dog dataset \cite{khosla2011novel}) have been developed for fine-grained object classification and recognition in 2D area, but large-scale 3D object datasets are still eagerly needed for fine-grained 3D shape classification.
Recently, several well-known large-scale  3D shape repositories, such as ShapeNet \cite{chang2015shapenet} and ModelNet \cite{shilane2004princeton}, have been developed for learning shape representations in various applications. While they contain large numbers of 3D objects from different categories, they are not well organized for fine-grained 3D shape classification. 
On the other hand, existing view-based methods for learning 3D shape representations are still suffering from capturing the fine-grained details from multiple views in a more comprehensive way, which is necessary for fine-grained classification of 3D shapes.
Intuitively, subtle and local differences are usually exposed in parts on the objects, so it is vital to leverage the parts in fine-grained 3D shape classification.

Previous multi-view methods such as MVCNN \cite{Su_2015_ICCV} and RotationNet \cite{kanezaki2018rotationnet} usually extract a feature from pixel-level information in each view first, and then aggregate the extracted view features into a global shape representation.
However, these methods do not capture part-level semantics from multiple views.
To address this problem, our recent work named Parts4Feature \cite{han2019parts4feature} utilized a region proposal network to detect generally semantic parts from multiple views and then learned to directly aggregate all generally semantic parts into a global shape representation.
However, there are still several unresolved issues in Parts4Feature, which limits its performance in fine-grained 3D shape classification.
First, Parts4Feature cannot capture the correlation of generally semantic parts in the same view, which makes it unable to filter out the meaningless generally semantic parts.
Second, Parts4Feature ignores view-level information such as the importance of each view and the spatial relationship among sequential views, which is important for learning the fine-grained 3D shape features.

To solve the above-mentioned issues, in this paper we propose a novel fine-grained 3D shape classification method, named FG3D-Net, as shown in Fig. \ref{fig:idea}, which leverages a hierarchical part-view attention aggregation module to capture the fine-grained features.
Similar to \cite{han2019parts4feature},  we first employ a region proposal neural network to detect generally semantic parts in each one of multiple views, which is considered to contain rich local details of 3D shapes.
By introducing the supervision information of bounding boxes from other 3D segmentation datasets, our FG3D-Net is able to explore the fine-grained details inside local parts.
Then, to aggregate all these extracted generally semantic parts, we leverage semantic information at different levels including part-level, view-level, and shape-level.
Specifically, we introduce part-level attention to highlight the important parts in each view and view-level attention to highlight discriminative views among all the views of the same object.
To take advantage of sequential input views as used in \cite{han2018seqviews2seqlabels},  we employ a Recurrent Neural Network (RNN) to encode the spatial relationship among views.
In order to eliminate the impact of the initial view in the RNN inputs, we integrate a global 3D shape feature with a max-pooling operation as \cite{Su_2015_ICCV}, which is invariant to the permutation of views.

In addition, we introduce a new fine-grained  3D  shape dataset that consists of three object categories including \emph{Airplane}, \emph{Car}, and \emph{Chair}, where dozens of subcategories are constructed in each category.
All 3D objects in the dataset are collected from several online repositories and  are organized  under  the  WordNet \cite{miller1995wordnet}  taxonomy.
Different from existing 3D datasets such as ShapeNet \cite{chang2015shapenet} and ModelNet \cite{shilane2004princeton}, our dataset is organized as a fine-grained 3D shape classification benchmark, where each 3D object is strictly assigned to one single subcategory in its category.
Our main contributions are summarized as follows.
\begin{itemize}
  \item We present a new fine-grained 3D shape dataset (named FG3D dataset) consisting of three categories including \emph{Airplane}, \emph{Car} and \emph{Chair}, which contains tens of thousands of 3D shapes with unique sub-category labels. This enables the learning of fine-grained features for fine-grained 3D shape classification.
  \item We propose a novel deep neural network named FG3D-Net to extract a global 3D shape representation that captures the fine-grained local details from generally semantic parts. FG3D-Net further includes a part-level and view-level attention mechanism to highlight the more semantic generally semantic parts in each view and the more distinctive views for each object, respectively. 
  \item We show that FG3D-Net outperforms state-of-the-art techniques in the fine-grained 3D shape classification task.
\end{itemize}

\begin{figure}
    \centering
    \includegraphics[width=8.5cm]{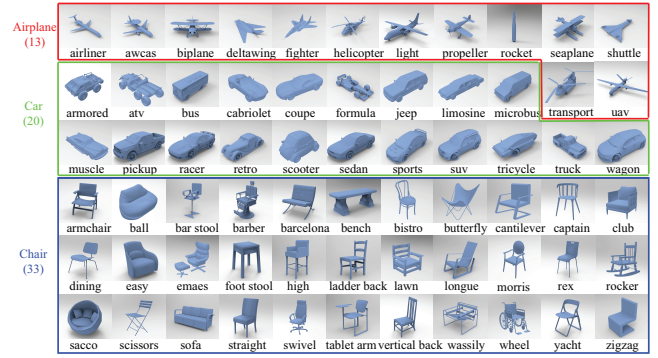}
    \caption{There are three shape categories in our fine-grained dataset including \emph{Airplane}, \emph{Car} and \emph{Chair}. 
    Specifically, 13 shape subcategories are included in the \emph{Airplane} category such as airliner, fighter and seaplane, 20 shape subcategories such as bus, jeep and scooter, are involved in the \emph{Car} category, and the \emph{Chair} category consists of 33 shape subcategories including bistro, captain, rocker, etc.}
    \label{fig:dataset}
\end{figure}

\section{Related Work}
\subsection{3D shape datasets}
3D shapes are widely used in various applications, such as robotics \cite{seok2012meshworm} and 3D modeling \cite{wang2018pixel2mesh}.
In recent years, 3D shape understanding has attracted a lot of research interest.
However, due to the inherent complexity of 3D shapes, 3D shape understanding is still a challenging problem in 3D computer vision.
Benefiting from deep learning models, deep neural network based methods have achieved significant performance in 3D shape recognition.
These methods require large-scale 3D datasets which are crucial for training deep neural networks and evaluating their performance.
Researchers have been working on building some large-scale repositories \cite{bogo2014faust,bronstein2008numerical,shilane2004princeton,koch2019abc}, which are widely adopted to evaluate deep neural networks in various applications.
With the development of web-scale datasets, ShapeNet \cite{chang2015shapenet} has collected a large-scale set of synthetic 3D CAD models from online open-sourced 3D repositories, including more than three million models and three thousand object categories.
Some other 3D repositories \cite{benhabiles2009framework,hu2012co,kalogerakis2010learning,wang2012active} were also proposed, which contain semantic labels for the segmented components of 3D shapes. 
Recently, PartNet \cite{mo2019partnet} provided more fine-grained part annotations to support fine-grained 3D shape segmentation tasks.
However, there is still no suitable 3D shape benchmark for the fine-grained 3D shape classification task so far.

To address this problem, some previous studies introduced several fine-grained image datasets and evaluated the results of fine-grained 3D shape classification.
FGVC-Aircraft \cite{maji2013fine} has collected ten thousand images of aircraft spanning 100 aircraft objects.
The Car dataset \cite{KrauseStarkDengFei-Fei_3DRR2013} contains 16,185 images from 196 subcategories of cars.
Unfortunately, previous 3D fine-grained datasets usually represent each 3D shape with a single 2D image, which can be regarded as the benchmarks of fine-grained 2D image classification.
Therefore, in this paper, we present a fine-grained 3D shape dataset containing 3D shapes represented by 3D meshes, where these shapes can be easily translated into other 3D data formats such as rendered views, point clouds and volumetric voxels.
In Fig. \ref{fig:dataset}, we show all the subcategories in our fine-grained dataset of three categories including airplanes, cars and chairs, respectively.
The construction of the fine-grained dataset will be illustrated in Section \uppercase\expandafter{\romannumeral3}. FG3D DATASET.
With the help of this proposed fine-grained 3D shape dataset, we can evaluate the ability of algorithms in capturing shape details under fine-grained 3D shape classification. 

\subsection{Fine-grained classification}
Fine-grained classification aims to classify many subcategories under a same basic-level category such as different cats, dogs or cars.
Due to the large intra-subcategory variance and the small inter-subcategory variance, fine-grained classification is a long standing problem in computer vision.
Recently, deep learning based methods have been widely applied to fine-grained image classification and achieved significant improvement over traditional methods.
From the perspective of fine-grained image classification, current methods can be summarized into three categories: 
(1) ensemble of networks based methods, 
(2) visual attention based methods, 
(3) part detection based methods.

Firstly, ensemble of networks based methods were proposed to learn different representations of images for better classification performance with multiple neural networks.
MGD \cite{wang2015multiple} trained a series of CNNs at multiple levels, which focus on different regions of interests in images.
B-CNN \cite{lin2015bilinear} was proposed with a bilinear CNN model, which jointly combined two CNN feature extractors.
Spatial Transformers \cite{jaderberg2015spatial} were proposed with a learnable model that consists of three parts including localization network, grid generator, and sampler.
The discriminative parts inside images were captured by four parallel spatial transformers on images and passed to the part description subnets.

Secondly, influenced by attention mechanisms, researchers also focus on searching discriminative parts dynamically, rather than dealing with images directly.
AFGC \cite{sermanet2014attention} employed an attention mechanism for fine-grained classification system, which utilized the information of multi-resolution corps to obtain the location and the object on the input images.

Finally, subtle and local differences are usually shown in discriminative parts of objects.
Therefore, discriminative part detection is very important for fine-grained shape classification.
The R-CNN family approaches \cite{ren2015faster} employed a popular strategy that first generates thousands of candidate proposals and then filters out these proposals with confidence scores and bounding box locations.
Zhang et al. \cite{zhang2014part} proposed to detect discriminative parts for fine-grained image classification and trained a classifier on the features of detected parts.
Recently, some studies \cite{simon2015neural,zhang2016weakly} focused on detecting discriminative parts under the weakly supervised setting, which means neither object nor part annotations are needed in both training and testing phases.
In \cite{simon2015neural}, part detectors were trained by finding constellations of neural activation patterns computed using convolutional neural networks.
Specifically,  the neural activation maps were computed as part detectors by using the outputs of a middle layer of CNN.
All these methods have been proposed to accomplish fine-grained classification of 2D images.
However, the fine-grained classification of 3D shapes has been rarely explored so far.
\begin{table*}
    \centering
    \caption{The statistics of our FG3D dataset which consists of 3 categories and 66 subcategories.}
    \begin{tabular}{ccccIccccIcccc} \shline
         Subcategory &Train &Test &\textbf{Total} &Subcategory &Train &Test &\textbf{Total} &Subcategory &Train &Test &\textbf{Total}     \\ \shline
         airliner &955 &100 &\textbf{1055}   &muscle &468 &100 &\textbf{568}                  &dining &614 &100 &\textbf{714}    \\ \hline
         awcas &24 &10 &\textbf{34}          &pickup &201 &50 &\textbf{251}                   &easy &990 &100 &\textbf{1090}       \\ \hline
         biplane &127 &50 &\textbf{177}      &racer &383 &100 &\textbf{483}                   &emaes &73 &40 &\textbf{113}     \\ \hline
         deltawing &186 &50 &\textbf{236}    &retro &102 &50 &\textbf{152}                    &foot stool &186 &50 &\textbf{236}       \\ \hline
         fighter &545 &100 &\textbf{645}     &scooter &25 &5 &\textbf{30}                     &high &256 &100 &\textbf{356}   \\ \hline
         helicopter &549 &100 &\textbf{649}  &sedan &2027 &100 &\textbf{2127}                 &ladder back &241 &50 &\textbf{291}      \\ \hline
         light &101 &30 &\textbf{131} 	    &sports &427 &100 &\textbf{527}                  &lawn &84 &30 &\textbf{114} \\ \hline
         propeller &288 &100 &\textbf{388}   &suv &209 &100 &\textbf{309}                     &longue  &665 &100 &\textbf{765} \\ \hline
         rocket &390 &100 &\textbf{490}      &tricycle &14 &5 &\textbf{19}                    &morris  &154 &50 &\textbf{204}  \\ \hline
         seaplane &35 &20 &\textbf{55}       &truck &131 &50 &\textbf{181 }                   &rex &407 &100 &\textbf{507}  \\ \hline
         shuttle &201 &50 &\textbf{251}      &wagon &427 &100 &\textbf{527 }                  &rocker    &121 &50 &\textbf{171}   \\ \hline \cline{5-8}
         transport &15 &7 &\textbf{22}       &\textbf{Car total} &\textbf{7010} &\textbf{1315} &\textbf{8325}   &sacco   &114 &50 &\textbf{164}  \\ \hline \cline{5-8}
         uav &25 &15 &\textbf{40}            &armchair &1578 &100 &\textbf{1678 }             &scissors   &65 &20 &\textbf{85}  \\ \hline \cline{1-4}
         \textbf{Airplane total} &\textbf{3441} &\textbf{732} &\textbf{4173} &ball &226 &100 &\textbf{326}      &sofa &307 &100 &\textbf{407}   \\ \hline \cline{1-4}
         armored &16 &5 &\textbf{21}         &bar &606 &100 &\textbf{706}                     &straight &1553 &100 &\textbf{1653} \\ \hline
         atv &77 &30 &\textbf{107}           &barber &20 &5 &\textbf{25}                      &swivel &777 &100 &\textbf{877}   \\ \hline
         bus &936 &100 &\textbf{1036}        &barcelona &35 &10 &\textbf{45}                  &tablet armed &51 &20 &\textbf{71}   \\ \hline
         cabriolet &477 &100 &\textbf{577}   &bench  &41 &20 &\textbf{61}                     &vertical back &748 &100 &\textbf{848}  \\ \hline
         coupe &528 &100 &\textbf{628}       &bistro &48 &20 &\textbf{68}                     &wassily &18 &5 &\textbf{23}  \\ \hline
         formula &62 &20 &\textbf{82}        &butterfly &10 &5 &\textbf{15}                   &wheel  &7 &5 &\textbf{12}  \\ \hline
         jeep &276 &100 &\textbf{376}        &cantilever &310 &100 &\textbf{410}              &yacht  &55 &20 &\textbf{75}  \\ \hline
         limousine &109 &50 &\textbf{159} &captain &217 &50 &\textbf{267}             &zigzag &89 &30 &\textbf{119}  \\ \hline \cline{9-12}
         microbus &115 &50 &\textbf{165}     &club &458 &100 &\textbf{558}                    &\textbf{Chair total} &\textbf{11124} &\textbf{1930} &\textbf{13054}  \\ \hline \cline{9-12}
    \end{tabular}
    \label{tab:fg3d_sta}
\end{table*}

\begin{table*}
    \centering
    \caption{The dataset comparison between our FG3D and ShapeNet.}
    \begin{tabular}{c|c|c|c|c} \hline
         Category &\#Subcategories &\#Overlap Subcategories  &\#Total Shapes &\#Overlap Shapes    \\ \hline
         Airplane (FG3D) &13 &\multirow{2}*{7} &4173 &\multirow{2}*{3064}	    \\  \cline{1-2} \cline{4-4}
         Airplane (ShapeNet)  &11 &~    &4045   &~     \\ \hline
         Car (FG3D) &20 &\multirow{2}*{6} &8235   &\multirow{2}*{3593}         \\ \cline{1-2} \cline{4-4}
         Car (ShapeNet) &18 &~ &4472   &~    \\ \hline
         Chair (FG3D) &33 &\multirow{2}*{20} &13054  &\multirow{2}*{6781} \\ \cline{1-2} \cline{4-4}
         Chair (ShapeNet)  &23 &~     &8591   &~       \\ \hline
    \end{tabular}
    \label{tab:compare_shapenet}
\end{table*}


To address this issue, we propose FG3D-Net to learn fine-grained global 3D shape features by capturing geometry details in generally semantic parts.

\subsection{Deep learning based methods for 3D shapes}
Benefiting from the advances in deep learning, deep learning based methods have achieved significant performance in 3D shape understanding tasks such as shape classification and recognition.
In general, current methods can be categorized into mesh-based, voxel-based, point cloud-based, and view-based deep learning methods.
To directly learn 3D features from 3D meshes, circle convolution \cite{han2016unsupervised} and mesh convolution \cite{han2016mesh} were proposed to learn local or global features.
Similar to images, voxels also have a regular structure that can be learned by deep learning models, such as CRBM \cite{wu20153d}, SeqXY2SeqZ \cite{han-eccv} and DSO-Net \cite{han2020reconstructing}, fully convolutional denoising autoencoders \cite{sharma2016vconv}, CNNs \cite{qi2016volumetric} and GANs \cite{wu2016learning}. 
These methods usually employ 3D convolution to better capture the contextual information inside local regions. 
Moreover, Tags2Parts \cite{muralikrishnan2018tags2parts} discovered semantic regions that strongly correlate with user-prescribed tags by learning from voxels using a novel U-Net.
As a series of pioneering work, PointNet \cite{qi2017pointnet} and PointNet++ \cite{qi2017pointnet++} inspired various supervised methods \cite{wen2020point2spatialcapsule,wen2020cf,Wen_2020_CVPR,han_icml,han2020shapecaptioner} to understand point clouds. 
Through self-reconstruction, FoldingNet \cite{yang2018foldingnet} and LatentGAN \cite{achlioptas2018learning,Han_2019_ICCV,han2019view} learned global features with different unsupervised strategies.
Similar to the light field descriptor (LFD), GIFT \cite{bai2017gift} measured the difference between two 3D shapes using their corresponding view feature sets. 
Moreover, pooling panorama views \cite{shi2015deeppano,sfikas2017exploiting} or rendered views \cite{su2015multi,han2019y2seq2seq,han2019parts4feature} are more widely used to learn global features.
Different improvements from camera trajectories \cite{johns2016pairwise}, view aggregation \cite{wang2017dominant,han2018seqviews2seqlabels,han2019view,han20193d2seqviews}, pose estimation \cite{Kanezaki_2018_CVPR} have been presented.
Parts4Feature \cite{han2019parts4feature} integrated a Region Proposal Network (RPN) to detect generally semantic parts in multiple views and then aggregated global shape feature from these generally semantic parts. 
However, it is still hard for current methods to fully explore the fine-grained details of 3D shapes in the fine-grained classification task.
In FG3D-Net, we introduce a hierarchical part-view attention aggregation strategy to extract more discriminative information from generally semantic parts.

\section{FG3D dataset}
To evaluate the performance in fine-grained 3D shape classification, we introduce a fine-grained 3D shape (FG3D) dataset.
Different from previous datasets such as ShapeNet \cite{Su_2015_ICCV}, FG3D aims to evaluate the fine-grained recognition of sub-categories within the same basic category, where 3D shapes may exhibit large intra-subcategory variance and small inter-subcategory variance.

A dataset for fine-grained 3D shape classification needs to fulfill six crucial properties \cite{koch2019abc,mo2019partnet}: 
(1) a large number of shapes for deep networks to capture statistically significant patterns; 
(2)  ground truth labels that enable to quantitatively evaluate the performance in a specific task; 
(3) unique fine-grained labels for each shape from dozens of subcategories under the same basic shape category; 
(4) convenient shape representation as input;
(5) 3D file format which deals with the challenges of 3D shape recognition; 
(6) expandability, i.e., make it easy for the collection to grow over time, to keep the dataset challenging as the performance of learning algorithms improves.
To build the FG3D dataset, we collect a large quantity of 3D shape files and classify each 3D shape into one unique subcategory, which strictly follows the above requirements.

Existing datasets are usually composed of 3D shapes from different basic categories such as ShapeNet \cite{chang2015shapenet} and ModelNet \cite{wu20153d}, but they do not satisfy the aforementioned property (3). 
Although ShapeNet includes multiple subcategories under each basic category, some 3D shapes have multiple or incorrect subcategory labels. This makes it impossible to use ShapeNet for fine-grained 3D shape classification.
In addition, the image-based datasets such as FGVC-Aircraft \cite{maji2013fine} and Car \cite{KrauseStarkDengFei-Fei_3DRR2013} do not satisfy property (5), since they do not contain any 3D shapes.

FG3D is complementary to current datasets by addressing fine-grained classification and satisfisying the properties (1)-(6). From the perspective of fine-grained visual classification, FG3D exhibits considerable variation within subcategories, but limited variation among different subcategories under each category. Hence, fine-grained classification on FG3D is a challenging task. To show the differences with ShapeNet, we compare FG3D with ShapeNet for the three categories in FG3D in TABLE \ref{tab:compare_shapenet}.
FG3D contains more 3D objects and subcategories to support the fine-grained classification task.

\begin{figure*}
    \centering
    \includegraphics[width=18cm]{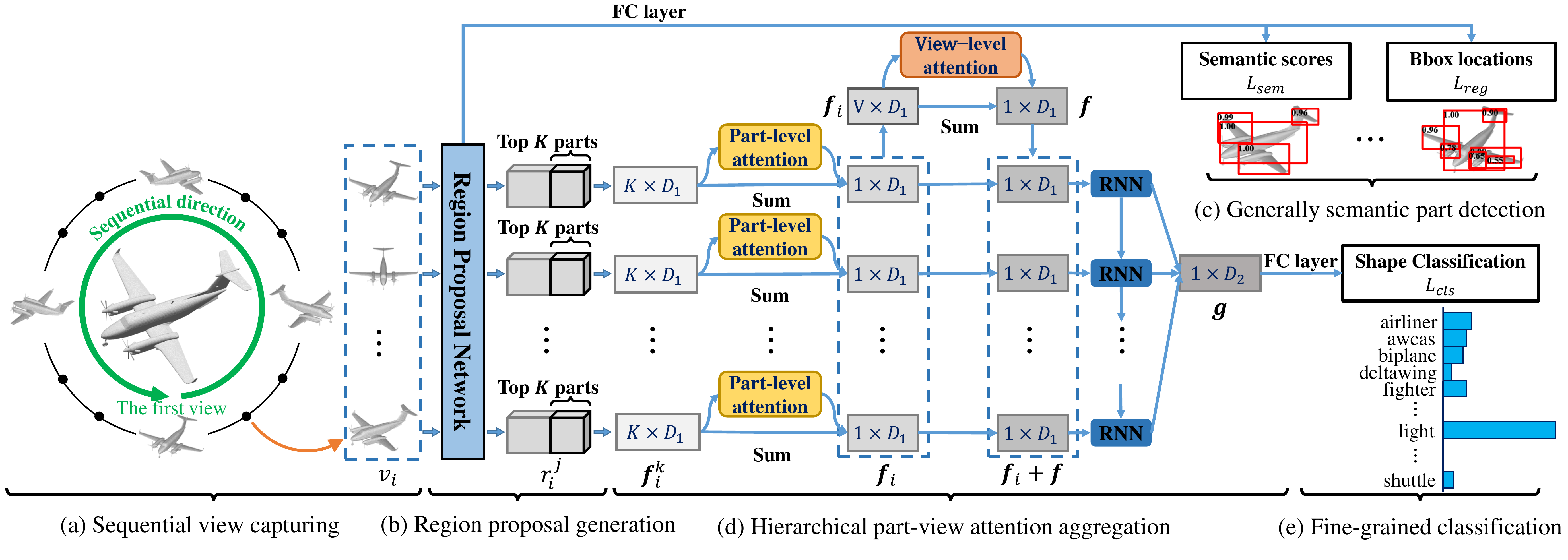}
    \caption{The framework of FG3D-Net.
    A sequence of views are first rendered from multiple viewpoints around the input 3D shape in the sequential view capturing module (a).
    In module (b), all the views are propagated into a region proposal network to generate region proposals and to compute the corresponding proposal features with RoI pooling  (b).
    Then, generally semantic parts in each view are detected by predicting the semantic scores and bounding box (bbox) locations with several FC layes in module (c).
    Next, in the hierarchical part-view attention aggregation module (d), the top \textbf{K} region proposals according to the semantic scores are selected to extract the global feature for the input 3D shape. 
    There are three different semantic levels in the feature aggregation mechanism, including part-level, view-level, and shape-level. 
    Two bilinear attention mechanisms are integrated to explore the correlation among features in different semantic levels. 
    In addition, an RNN layer is applied to enhance the correlation of view features by taking advantage of the sequential input views.
    Finally, the global shape representation is extracted by a max-pooling layer, which is applied to the fine-grained classification of 3D shapes in the module (e).
    }
    \label{fig:framework}
\end{figure*}
As shown in TABLE~\ref{tab:fg3d_sta}, FG3D consists of three basic categories including \textit{Airplane}, \textit{Car} and \textit{Chair}, which contain 3,441 shapes in 13 subcategories, 8,235 shapes in 20 subcategories, and 13,054 shapes in 33  subcategories, respectively.
We represent each 3D shape by an object format file (.off) with polygonal surface geometry. 
One can easily convert the .off files into other shape representations, such as rendered views, voxels and point clouds.
All shapes in FG3D are collected from multiple online repositories including 3D Warehouse \cite{goldfeder2008autotagging}, Yobi3D \cite{yobi3d2019shapes} and ShapeNet \cite{chang2015shapenet}, which contain a massive collection of CAD shapes that are publicly available for research purpose.
By collecting 3D shapes over a period of two months, we obtained a collection of more than 20K 3D shapes in three shape categories.
We organized these 3D shapes using the WordNet \cite{miller1995wordnet} noun ``synsets" (synonym sets).
WordNet provides a broad and deep taxonomy with over 80K distinct synsets representing distinct noun concepts. 
This taxonomy has been utilized by ImageNet \cite{deng2009imagenet} and ShapeNet \cite{chang2015shapenet} to formulate the object subcategories.
In our dataset, we also introduce the taxonomy into the collection of 3D shapes, as shown in Fig. \ref{fig:dataset}.

For evaluation, we split the shapes in each categories into training and testing sets.
Specifically, the 3D shapes in airplane are split into 3,441 for training and 732 for testing.
The cars category contains 7,010 shapes for training and 1,315 shapes for testing.
The chairs category contains 11,124 shapes for training and 1,930 shapes for testing.


\section{FG3D-Net}
\subsection{Overview}
As shown in Fig. \ref{fig:framework}, The framework of FG3D-Net consists of five main modules including (a) sequential view capturing, (b) region proposal generation, (c) generally semantic part detection, (d) hierarchical part-view attention aggregation and (e) fine-grained classification.
In particular, the modules (b) and (c) compose a region proposal network (RPN) and cooperate to complete the detection of Generally Semantic Parts (GSPs) from multiple rendered views, which are pre-trained under several part segmentation benchmarks.
To construct the region proposal network, we follow the similar strategy as in Parts4Feature \cite{han2019parts4feature} which also detects the GSPs from multiple views.
By introducing the part information from other segmentation data, our FG3D-Net can integrates the fine-grained details inside local parts.

For each input 3D shape $\mathcal{M}$ from the training set, a view sequence $\bm{v}$ is first obtained by rendering $V$ views $\{v_i\}$ around $\mathcal{M}$, such that $\bm{v} = [v_1,\cdots,v_i,\cdots,v_V]$ and $i \in [ 1,V]$, as shown in the module (a) of Fig. \ref{fig:framework}.
Then, a shared convolutional neural network (e.g., VGG19 \cite{simonyan2014very}) abstracts all the views into feature maps of the high-dimensional feature space.
By applying a sliding window on the feature maps, numerous region proposals $\{r_i^j\}$ are calculated for each view $v_i$, where the corresponding proposal features $\{\bm{c}_i^j\}$ are extracted by a RoI (Region-of-Interest) pooling layer and $j \in [1, N]$.
With proposal features $\bm{c}_i^j$, module (c) learns to predict both the semantic scores and bounding box locations of GSPs within multiple views.
Finally, according to the semantic scores, part features $\{\bm{c}_i^k\}$ of the top $K$ region proposals $\{r_i^k\}$ in each view $v_i$ of $\bm{v}$ are selected for extracting the global shape feature $\bm{f}$ of $\mathcal{M}$ with the module (d). 
The global shape feature $\bm{f}$ is propagated through a Fully Connected (FC) layer to  provide the classification probability $\bm{p}$ for fine-grained 3D shape classification.
\begin{figure}
    \centering
    \includegraphics[width=7cm]{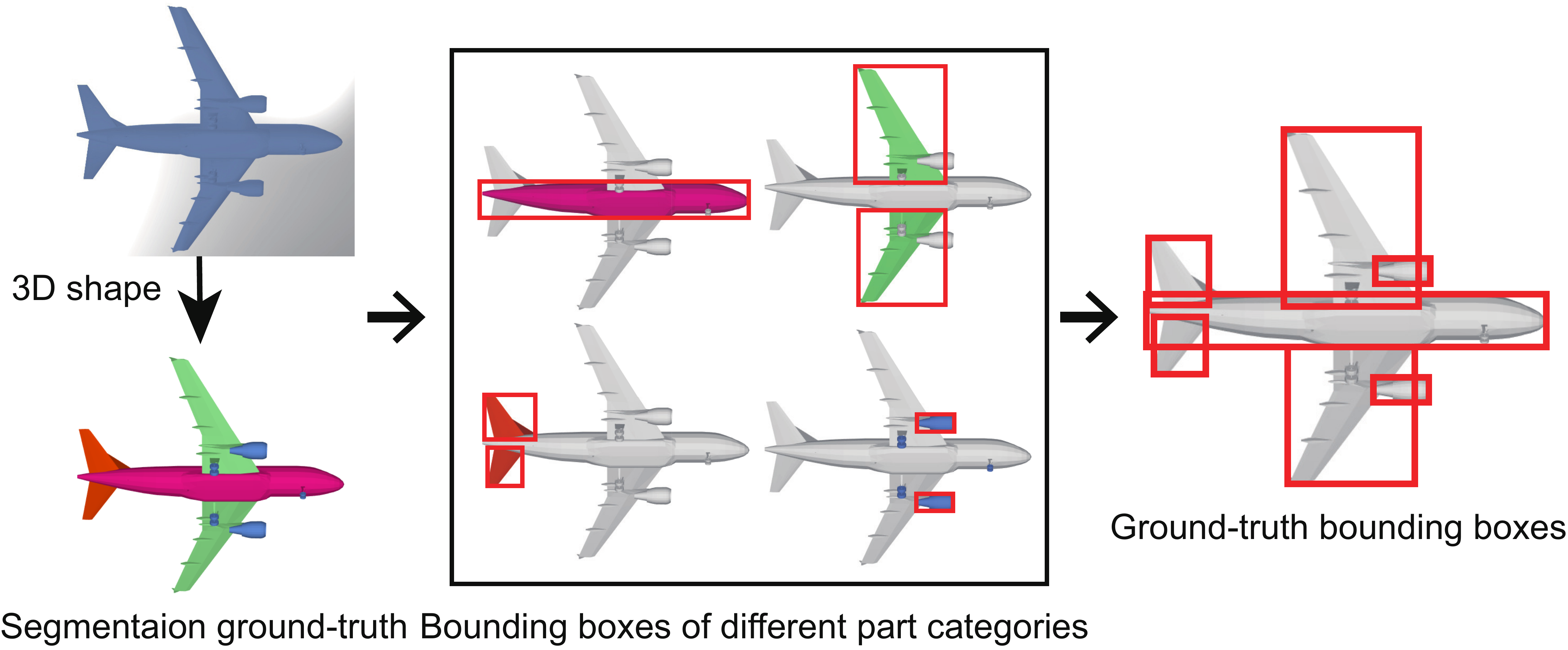}
    \caption{The generation of ground-truth generally semantic parts from segmentation. According to the segmentation ground-truths, we first individually render each 3D part into 2D views with different colors. Then, we extract the bounding boxes of the colored parts using image processing. Finally, we obtain the ground-truth bounding-boxes of each 3D part in the 2D views.}
    \label{fig:segement}
\end{figure}

\subsection{Generally semantic parts (GSPs)}
A generally semantic part (GSP) in FG3D-Net indicates a local part in rendered views of a 3D shape. GSPs do not distinguish between different semantic part classes or shape categories, such as the wings of airplanes or the wheels of cars.
Different from semantic parts in 3D shape segmentation, a generally semantic part in FG3D-Net represents a local visual part in 2D views, rather than shape parts in 3D.
By learning the shape representation form GSPs, our method exploits the fine-grained details of multiple views at three different feature abstration levels, including part-level, view-level, and shape-level.

In the pre-training stage of GSPs detection, all ground-truth GSPs are generated from several 3D shape segmentation benchmarks.
By learning the GSPs from other 3D segmentation datasets, we are able to detect the GSPs in multiple views under our fine-grained classification benchmark without requiring ground-truth GSP supervision in the FG3D dataset.
Specifically, we use three 3D shape segmentation benchmarks including ShapeNetCore \cite{yi2016scalable}, Labeled-PSB \cite{chen2009benchmark,kalogerakis2010learning}, and COSEG \cite{chen2009benchmark} to construct the generally semantic part detection benchmark and provide ground-truth GSPs. 
Here, we follow \cite{kalogerakis20173d} to split the 3D shapes into training and testing set.

Fig. \ref{fig:segement} shows the pipeline to generate the ground-truth bounding boxes for GSPs.
Based on the segmentation ground-truth in 3D, we represent each segmented 3D part with a different color.
To extract the bounding box for each 3D part, we render each colored 3D part into 2D images. 
With a simple image processing step, we apply a region property measurement function to calculate the bounding box of the colored parts.
To reduce the impact of data noise, we apply a data cleaning step to eliminate the influence of some small parts whose bounding boxes are smaller than 0.45 of the max bounding box in the same part category.
In addition, another benefit of using GSPs is eliminating the impact of some incorrect segmentations, where 3D parts may have wrong segmentation labels in the datasets.
So far, we have obtained the bounding boxes of the ground truth GSPs within multiple views, where the ground-truth GSPs are applied to train modules (b) and (c) for generally semantic part detection, as shown in Fig. \ref{fig:framework}.


\subsection{Region proposal network}
A Region Proposal Network (RPN) takes an image as input and outputs a set of rectangular object proposals, each with an objectness score. 
As shown in Fig. \ref{fig:framework}, the modules (a) and (b) of our FG3D-Net comopse a RPN, which is adjusted for detecting GSPs from the multiple views of 3D shapes.
Similar to \cite{ren2015faster}, the RPN detects a set of generally semantic part proposals with corrsponding semantic scores for each view.
Specifically, in module (b), a large number of region proposal candidates $\{r_i^j\}$ are first generated for each view $v_i$ and the corresponding proposal feature $\{\bm{c}_i^j\}$ are calculated by a Region-of-Interest (RoI) pooling layer.
To complete the regeion proposal generation, there are a Convolutional Neural Network (CNN) layer, a region proposal calculation layer and a RoI pooling layer in the module (b).
Then, in module (c), all the proposal features $\{\bm{c}_i^j\}$ are applied to predict both the semantic scores and locations of GSPs with several stacked Fully Connected (FC) layers.
According to the predicted semantic scores, the features $\{\bm{c}_i^k\}$ of the top $K$ proposals are selected for hierarchically learning the 3D shape representations in module (d).

To review the details of the RPN, a Convolutional Neural Network (CNN) first abstracts multiple input views into the feature maps.
The CNN layer is modified from a VGG-19 network proposed in \cite{simonyan2014very}, and it produces a feature map $\bm{c}_i$ for each view $v_i$. 
Secondly,  a region proposal layer calculates the region proposals $\{r_i^j\}$ by sliding a small window over the corresponding feature map $\bm{c}_i$.
At each sliding-window location, which is centered at each pixel $\bm{c}_i$, a region $r_i^j$ is proposed by regressing its location $t_D$ and predicting a semantic probability $p_D$ with an anchor. 
The location $t_D$ is a four dimensional vector representing center coordinates, width and height of the part bounding box. 
All of the above modifications are adjusted from previous Faster-RCNN \cite{girshick2014rich}.

To train RPN for predicting the semantic scores $p_D$ of GSPs, we assign a binary label to each region proposal $r_i^j$ to indicate whether $r_i^j$ is a GSP. 
Specifically, we assign a positive label if the IoU (Intersection-over-Union) overlap between $r_i^j$ and any ground-truth GSP in $v_i$ is higher than a threshold $S_D$.
Note that a single ground-truth box may assign positive labels to multiple anchors; otherwise, we use a negative label.
In each view $v_i$, we apply RoI pooling over region proposal location $t_D$ on feature maps $\bm{c}_i$.
Hence, the features $\{\bm{c}_i^j\}$ of all $N$ region proposals $\{r_i^j\}$ are high dimensional vectors, which we forward to the generally semantic part detection module as shown in Fig. \ref{fig:framework}. 
To reduce the computational cost, we represent each generally semantic part feature in the top $K$ proposals with a high dimensional feature vector $\bm{f}_i^k$ that is calculated by $\bm{f}_i^k = MAX(\{\bm{c}_i^k\})$.
Therefore, we finally select the features $\{\bm{f}_i^k\}$ of the top $K$ region proposals $\{r_i^k\}$ according to corrsponding semantic scores $p_D$, which are propagated to the hierarchical part-view attention aggregation module.

With the above definitions, an objective function is employed to optimize the part detection following the multi-task loss in Faster-RCNN \cite{ren2015faster}.
Denote that the ground-truth semantic scores and box locations of positive or negative samples in the RPN are $p^*$ and $t^*$, respectively.
The loss function is formulated as
\begin{equation}
    L(p_D, p^*, t_D, t^*) = L_{sem}(p_D, p^*) + \lambda L_{reg}(t_D, t^*),
    \label{Eq:deetction}
\end{equation}
where $L_{sem}$ measures the accuracy in terms of semantic scores by calculating the cross-entropy loss with part labels, while $L_{reg}$ measures the accuracy of location in terms of the $L_1$ distance \cite{ren2015faster}.
In the experiment, $\lambda$ works well with a value of 1.
By introducing the architecture of RPN \cite{ren2015faster}, FG3D-Net has the special ability to detect GSPs from multipe views by being pre-trained under other segmentation benchmarks.

\subsection{Hierarchical part-view attention aggregation}
In order to extract shape features with fine-grained distinction, the hierarchical part-view attention aggregation module (d) is the key component of FG3D-Net, as shown in Fig. \ref{fig:framework}, which hierarchically learns the global representation $\bm{f}$ of the input 3D shape from the features $\{\bm{f}_i^k\}$ of generally semantic parts.
To propagate the features from low-level to high-level, it is important to preserve the detailed information in different levels.
Thus, in module (d), there are three different semantic levels including part-level, view-level and shape-level, where the fine-grained relationship among features are explored.
To aggregate shape features from low to high, some special designs are introduced including part-level attention, view-level attention and view feature enhancement.
\begin{figure}
    \centering
    \includegraphics[height=2.5cm]{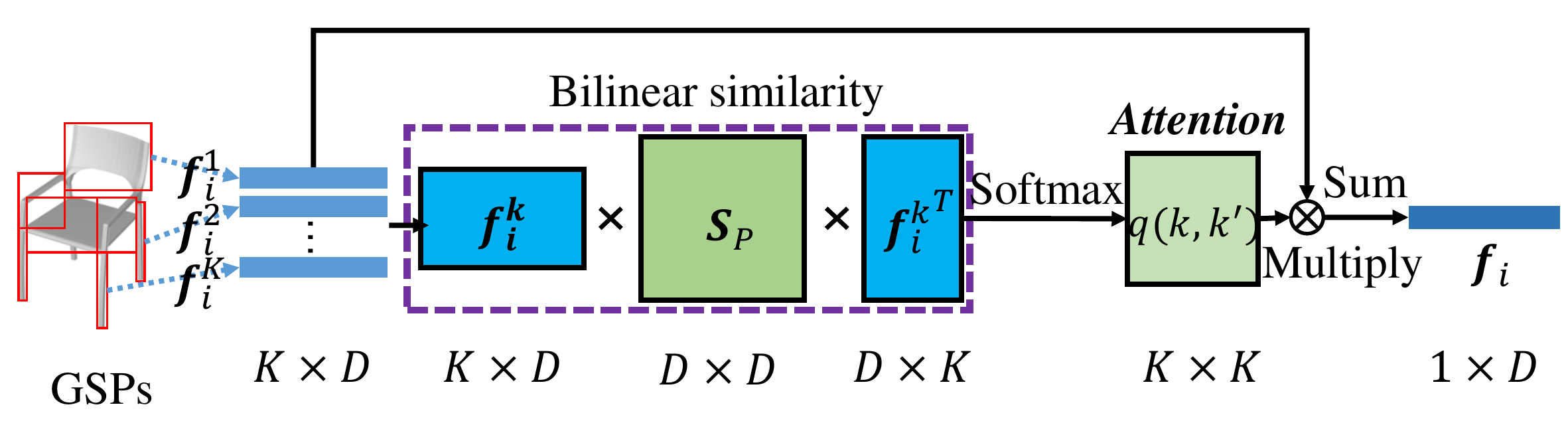}
    \caption{The demonstration of part-level attention. By calculating the attention values among generally semantic parts, we aggregate $K$ parts features $\{\bm{f}_i^k\}$ in the same view into a view feature $\bm{f}_i$.
    Firstly, we compute the bilinear similarity score $q(k,k^{'})$ between $\bm{f}_i^k$ and $\bm{f}_i^{k^{'}}$ with a softmax layer.
    Then, we multiply the scores $q(k,k^{'})$ to the input part features to highlight the discriminative features.
    Here, $\otimes$ indicates multiplication.
    Finally, a weighted sum is applied to aggregate the highlighted parts features into a view feature $\bm{f}_i$. 
    }
    \label{fig:part_attention}
\end{figure}

\subsubsection{Part-level attention}
The target of part-level attention is to aggregate the features of GSPs detected from the same view.
As shown in Fig. \ref{fig:part_attention}, we first select the top $K$ GSP features $\{\bm{f}_i^k\}$ of view $v_i$ to extract the corresponding view feature $\bm{f}_i$.
Traditional approaches sucha as \cite{Su_2015_ICCV} usually aggregate multiple features by simple max-pooling or mean-pooling operations.
However, simple pooling operation suffers from the content information loss within generally semantic parts.
As summarized by some previous methods \cite{zheng2017learning,wang2017multi,guo2019aligned,ge2019weakly} for fine-grained classification of 2D images, detailed information about local parts usually determines the discriminability of learned object features.
To resolve this issue, we propose a bilinear similarity attention to aggregate generally semantic part features $\{\bm{f}_i^k\}$ into view feature $\bm{f}_i$ for view $v_i$ in the view sequence $\bm{v}$.
The part-level attention is designed to take advantage of the relationships among part features to facilitate the feature aggregation procedure.
Specifically, a shared matrix $\bm{S}_p$ is learned to evaluate the mutual correlations among the $K$ generally semantic part features $\{\bm{f}_i^k\}$ of view $v_i$.
Here the learnable matrix $\bm{S}_p$ is shared across all views, which aims to capture the general patterns among GSPs in each one of multiple views.
In addition, the learning of $\bm{S}_p$ can also explore the local patterns of GSPs from different 3D shapes.
By applying an attention value to each GSP, the part-level attention can highlight the important GSPs in each view to facilitate feature aggregation.
\begin{figure}
    \centering
    \includegraphics[height=2.5cm]{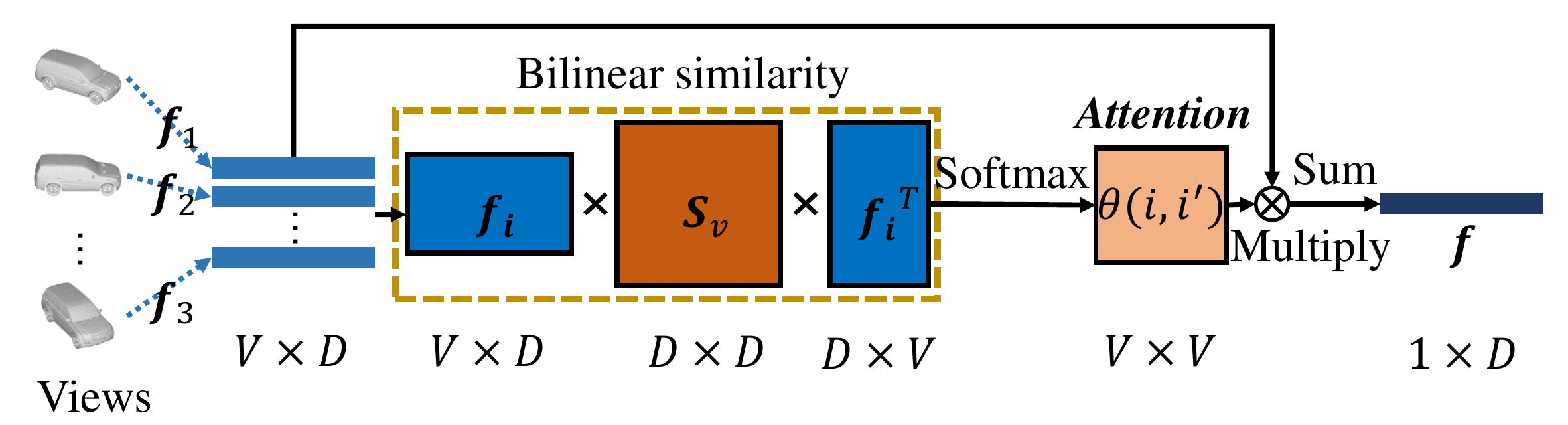}
    \caption{The demonstration of view-level attention. 
    Similar to the part-level attention, we calculate the bilinear similarity scores $\theta(i,i^{'})$ among the view features from the same 3D object to aggregate view features $\{\bm{f}_i\}$ into a shape feature $\bm{f}$.
    By applying a multiply operation $\otimes$ and a weighted sum operation, the global feature $\bm{f}$ is obtained by considering the importance of different view features.
    }
    \label{fig:view_attention}
\end{figure}

Given the GSP features $\{\bm{f}_i^k\}$ of view $v_i$, the corresponding context of $\bm{f}_i^k$ is formed by
\begin{equation}
 \bm{R}^{context}_{i,k}=\{\bm{f}_i^k\}, k \in [1,K].
\end{equation}
For each candidate $\bm{f}_i^k$, there is a score $q(k,k^{'})$ measuring the similarity between $\bm{f}_i^k$ and $\bm{f}_i^{k^{'}}$ as follows
\begin{equation}
    q(k,k^{'}) = \frac{exp({\bm{f}_i^k} \bm{S}_p {\bm{f}_i^{k^{'}}}^T)}{\sum_{n=1}^{K} exp({\bm{f}_i^k} \bm{S}_p {\bm{f}_i^n}^T)},
\end{equation}
where $\bm{S}_p$ is the learnable matrix. 
All $q(k,k^{'})$ form the bilinear similarity attention matrix with a size of $K \times K$, which represents the correlation among the $K$ GSPs.
Without extra selection of highly related ones, the context $\bm{e}_i^k$ fused for $\bm{f}_i^k$ is a weighted average over all the candidate parts, as denoted by
\begin{equation}
    \bm{e}_i^k = \sum_{{k^{'}}=1}^{K} q(k,k^{'}) \bm{f}_i^{k^{'}}.
\end{equation}
With the context vector of GSPs in view $v_i$, the view feature $\bm{f}_i$ can be calculated by
\begin{equation}
    \bm{f}_i = \sum_{k=1}^K \bm{e}_i^k ,
\end{equation}
where the view features $\{\bm{f}_i\}$ are propagated to both the view-level attention and the view feature enhancement module.

\subsubsection{View-level attention}
In the view-level, we apply a similar strategy to effectively aggregate view features $\bm{f}_i$ into 3D global features $\bm{f}$ as depicted in Fig. \ref{fig:view_attention}, which contains the global information of an entire 3D shape from multiple views.
To learn the attention value of view features from the same 3D object, one learnable parameter matrix $\bm{S}_v$ is also learned to capture the correlation among views.
Therefore, by calculating the bilinear similarity among view features, the attention values are captured, which can also leverage the importance of views in the feature aggregation.  
\begin{figure}
    \centering
    \includegraphics[height=2.8cm]{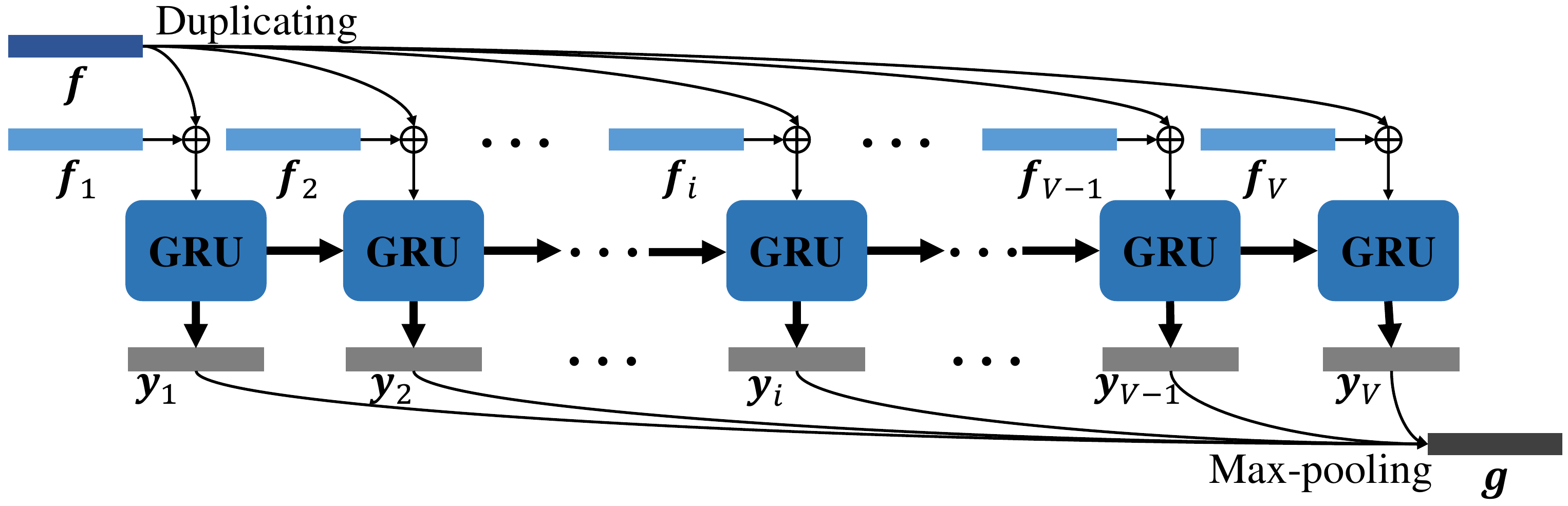}
    \caption{The demonstration of the view feature enhancement. 
    To enhance the view features, we first add the global shape feature $\bm{f}$ to $\bm{f}_i$ by duplicating, where $\oplus$ indicates the sum operation.
    Then, we adopt a GRU (RNN) to capture the spatial correlation among views.
    Therefore, we obtain the enhanced view features $\{\bm{y}_i\}$ which are aggregated into the final global shape representation $\bm{g}$ by a max-pooling layer.}
    \label{fig:vfe}
\end{figure}

Given the view features $\{\bm{f}_1,\bm{f}_2,\cdots,\bm{f}_i,\cdots,\bm{f}_V \}$, a similarity score $\theta(i,i^{'})$ measuring the similarity between $\bm{f}_i$ and $\bm{f}_{i^{'}}$ is computed as
\begin{equation}
    \theta(i,i^{'}) = \frac{exp(\bm{f}_i \bm{S}_v {\bm{f}_{i^{'}}}^T)}{\sum_{l=1}^V exp(\bm{f}_i \bm{S}_v {\bm{f}_{l}}^T)},
\end{equation}
where $\bm{S}_v$ is also a learnable matrix  to capture the correlation among views. 
The learned $\theta(i,i^{'})$ are the bilinear similarity attention matrix with a size of $V \times V$.
According to the learned similarity scores, the context $\bm{\bar{e}}_i$ of view $v_i$ is fused from $\bm{f}_i$ with a weighted average over the views
\begin{equation}
\bm{\bar{e}}_i = \sum_{{i^{'}}=1}^{V} \theta(i,i^{'}) \bm{f}_{i^{'}}.
\end{equation}
Finally, the 3D global feature $\bm{f}$ is computed as
\begin{equation}
    \bm{f} = \sum_{i=1}^{V} \bm{\bar{e}}_i.
\end{equation}

Through the hierarchical part-view attention mechanisms, we obtain the global shape representation $\bm{f}$, which can be applied for 3D shape recognition.
However, the spatial correlation among sequential views is not fully explored.
To take advantage of the prior information of the view sequence $\bm{v}$, we additionally utilize a recurrent neural network to enhance current 3D global feature $\bm{f}$.

\subsubsection{View feature enhancement}
An important prior information of sequential input views $\bm{v}$ is the spatial information among them, where the rendering viewpoints are continuously distributed around the shape in 3D space as shown in Fig. \ref{fig:framework}.
To benefit from the powerful ability of learning from sequential data, FG3D-Net employs a Recurrent Neural Network (RNN) to learn the enhanced view features $\{\bm{y}_i\}$ from previous view features $\{\bm{f}_i\}$ and global feature $\bm{f}$.

With the view features $\{ \bm{f}_1, \bm{f}_2, \cdots, \bm{f}_i, \cdots, \bm{f}_V \}$ aggregated from the part level, we first integrate the global shape information by
\begin{equation}
 \bm{f}^{'}_i = \bm{f}_i + \bm{f}.
\end{equation}
The integrated features $\bm{f}^{'}_i$ form a feature sequence $\bm{f}^{'}_{\bm{v}} = \{ \bm{f}^{'}_1, \bm{f}^{'}_2,\cdots,\bm{f}^{'}_i,\cdots,\bm{f}^{'}_V \}$.
A RNN takes the sequential view features as input and captures the spatial correlation among views.
The RNN consists of a hidden state $\bm{h}$ and an optional output $\bm{y}$, which operates on the view feature sequence $\bm{f}^{'}_{\bm{v}}$.
Here each item $\bm{f}^{'}_i$  is a 512-dimensional feature vector and the length of $\bm{f}^{'}_{\bm{v}}$ is $V$ which is also the number of steps in the RNN.
At each time step $t\in[1,V]$, the hidden state $\bm{h}_t$ of the RNN is updated by
\begin{equation}
    \bm{h}_t = GRU(\bm{h}_{t-1}, \bm{f}^{'}_t),
\end{equation}
where $GRU$ is a non-linear activation function named gated recurrent unit \cite{cho-etal-2014-learning}.

A RNN can learn the probability distribution over a sequence by being trained to predict the next item in the sequence. Similarly, at time step $t$, the output $\bm{y}_t$ of the RNN can be represented as
\begin{equation}
 \bm{y}_t = \bm{W}_a \bm{h}_t,
\end{equation}
where $\bm{W}_a$ is a learnable weight matrix.
After forwarding the entire input feature sequence, as shown in Fig. \ref{fig:vfe}, the output sequence $\{ \bm{y}_1, \bm{y}_2, \cdots, \bm{y}_i, \cdots, \bm{y}_V \}$ is acquired,
which contains the content information and the spatial information of the entire view feature sequence.
To avoid a dependency on the choice of the initial input view feature in the RNN, we learn a more general global shape feature $\bm{g}$ by adopting a max-pooling layer,
\begin{equation}
    \bm{g} = \mathop{\max}\limits_{i \in [1,V]}\{\bm{y}_i\}.
\end{equation}
Through the view feature enhancement, a global representation $\bm{g}$ is extracted from GSPs in multiple views.
Following a FC layer, $\bm{g}$ is applied to predict 3D shape labels with the cross-entropy loss function, where the softmax function outputs the classification probabilities $\bm{p}$.
Suppose that there are $C$ shape subcategories of an individual category in the classification, so each probability $\bm{p}(c)$ can be defined as
\begin{equation}
    \bm{p}(c) = \frac{exp({\bm{W}_c\bm{g} + a_c})}{\sum_{c^{'} \in [1, C]} exp(\bm{W}_{c^{'}}\bm{g} + a_{c^{'}})},
\end{equation}
where $\bm{W}$ and $a$ are the weights of the FC layer to be learned. The objective loss function is the cross-entropy between the predicted probability $\bm{p}$ and the corresponding ground truth $\bm{p}^{'}$,
\begin{equation}
    L{(\bm{p}, \bm{p^{'}})} = - \sum_{c \in [1, C]} \bm{p^{'}}(c) log(\bm{p}(c)).
    \label{Eq:classify}
\end{equation}

\subsection{Training}
There are two different tasks in our FG3D-Net including generally semantic part detection and fine-grained shape classifcation.
To leverage the performance of the two tasks, we use an alternating strategy to train FG3D-Net.
First, we train the region proposal network to detect generally semantic parts inside shape views under the processed segmentation benchmarks.
Then, while fixing the parameters in the region proposal network, we only update parameters inside the hierarchical part-view attention aggregation branch, as shown in Fig. \ref{fig:idea}.
By repeating the above two steps, we can apply our FG3D-Net to extract discriminative shape features for 3D shapes for fine-grained 3D shape classification. 
Therefore, the optimization target $L_{total}$ of FG3D-Net consists of a sum of two parts,  
\begin{equation}
    L_{total} = L(p_D, p^*, t_D, t^*) + \psi L{(\bm{p}, \bm{p^{'}})},
\end{equation}
where $\psi$ is a hyperparameter to balance loss terms.

\section{Experiments}
In this section, we conduct comprehensive experiments to validate FG3D-Net for fine-grained classification of 3D shapes.
We first explore how some key parameters affect the performance of FG3D-Net and then discuss results of an ablation study to justify our architecture.
Finally, under our FD3D dataset, we compare the fine-grained shape classification performance of FG3D-Net with state-of-the-art methods with different 3D shape representations.
\begin{table}[htp]
    \centering
    \caption{The key parameters in FG3D-Net.}
    \begin{tabular}{cc} \hline
    $V$      &the number of input views  \\ \hline
    $K$      &the number of GSPs in each view \\ \hline
    CT     &the cell type of RNN \\ \hline
    Dim    &the  dimension of RNN hidden state\\ \hline
    $S_D$  &the threshold of positive GSP\\ \hline
    $N$     &the number of proposal candidates \\ \hline
    $\bm{S}_p, \bm{S}_v$ &the learnable bilinear matrices \\ \hline
    $\lambda, \phi$   &the ratio of loss function terms \\ \hline
    \end{tabular}
    \label{tab:parameters}
\end{table}
\subsection{Network configuration}
As shown in TABLE \ref{tab:parameters}, we illustrate the key parameters in our FG3D-Net.
In the experiment, we apply default $V=12$ views of each 3D shape $\mathcal{M}$, where the viewpoints are uniformly distributed around the shape and each view size is $224 \times 224 \times 3$.
After the generally semantic part detection with $S_D = 0.7$, the features of top $K=20$ parts in each view are propagated to the hierarchical part-view attention aggregation module, as shown in Fig. \ref{fig:framework}.
Following some parameter settings as in Faster RCNN \cite{ren2015faster}, the size of the abstracted feature map $\bm{c}_i$ is $12 \times 12 \times 512$ for each view. 
To generate anchors on the feature map, We apply 6 scales and 3 aspect ratios to yield $6 \times 3 = 18$ anchors at each  pixel,  which ensures a wide coverage of sizes to accommodate region proposals for GSPs that may be partially occluded in some views.  
The 6  scales  relative  to  the  size  ofthe  views  are [1,2,4,8,16,32],  and  the  3  aspect  ratios  are 1 : 1, 1 : 2,  and 2 : 1.  
In  general,  all  anchors  lead  to $N = 2592 = 12 \times 12 \times 18$ proposal candidates $\{ r^j_i \}$ in each view $v_i$. 
And each region proposal feature $\bm{c}^j_i$ extracted by RoI pooling are with size of $512 \times 7 \times 7$.
In the part-view attention mechanisms, the learnable bilinear matrices $\bm{S}_p$, $\bm{S}_v$ are with size $512 \times 512$, and the demensions of all features are 512.
In the view feature enhancement module, the GRU cell \cite{cho-etal-2014-learning} is adopted in our RNN and the dimension of the hidden state is initialized with 4,096.
And the hyperparameters $\lambda$, $\psi$ are set to 1 in the loss function.
For all experiments, we train our network on a NVIDIA GTX 1,080Ti GPU using ADAM optimizer with an initial learning rate of 0.00001 and a batch size of 1.

\subsection{Parameter setting}
There are several important hyperparameters settings in FG3D-Net.
To investigate the influence of these hyperparameters, we performed comparisons with different settings under the category of airplane, which contains 4,173 3D shapes from 13 subcategories.
We first explore the number of input views $V$, which determines the coverage of input 3D shapes from different view angles.
In this experiment, we keep all other hyperparameters fixed and modify the number of rendered views $V$ from 3 to 20.
All views are obtained by rendering around each 3D shape as shown in MVCNN \cite{su2015multi}.
The results are listed in TABLE \ref{tab:V_comp}, which shows the accuracy trend with increasing the number of views.
With $V=12$ or $V=20$ views as input, FG3D-Net reaches the similar instance accuracy of $93.99\%$.
This is because 12 views have already covered most of the details of 3D shapes.
In addition, 12 views have been widly used as input of view-based methods, such as MVCNN \cite{Su_2015_ICCV}, SeqViews2SeqLabels \cite{han2018seqviews2seqlabels}, 3D2SeqViews \cite{han20193d2seqviews} and Parts4Feature\cite{han2019parts4feature}.
For fair comparison with these methods and leveraging the computational complexity of the whole network, we also adopt 12 views as input in our FG3D-Net.
\begin{table}[htp]
\centering
\caption{The effect of the view number $V$ on the performance in FG3D-Net.}
\begin{tabular}{ccccc} \hline
$V$ &3&6&12 &20\\ \hline
Acc (\%) & 88.39 & 92.90 & \textbf{93.99} &\textbf{93.99}\\
\hline\end{tabular}
\label{tab:V_comp}
\end{table}

In the following experiments, we keep the number of input views $V=12$.
In TABLE \ref{tab:K_comp}, we study the effect of the number $K$ of generally semantic parts in the hierarchical part-view attention aggregation module in Fig. \ref{fig:framework} (d), where $K$ ranges from 5 to 40.
The best accuracy $93.99 \%$ is reached at $K=20$, which can achieve a better coverage of the fine-grained details of 3D shapes.
\begin{table}[htp]
\centering
\caption{The effect of the number of generally semantic parts $K$ on the performance in FG3D-Net.}
\begin{tabular}{cccccc} \hline
$K$ &5&10&20&30&40\\ \hline
Acc (\%) & 92.90 & 93.31 & \textbf{93.99}  & 93.58  & 93.44 \\
\hline\end{tabular}
\label{tab:K_comp}
\end{table}

To investigate the effect of the RNN cell type (CT) in our \textit{view feature enhancement} step, we show the results with different RNN cells in TABLE \ref{tab:CT_comp}.
We observe that the GRU cell outperforms other RNN cells such as BasicRNN, LSTM and BidirectionRNN.
In particular, in the BidirectionRNN cell, there are two GRU cells of different directions.
\begin{table}[htp]
\centering
\caption{The effect of RNN Cell types (CTs) on the performance in FG3D-Net.}
\begin{tabular}{ccccc} \hline
CT & BasicRNN & LSTM & BidirectionRNN & GRU\\ \hline
Acc (\%) & 91.80 & 92.76 & 93.44 & \textbf{93.99} \\
\hline\end{tabular}
\label{tab:CT_comp}
\end{table}

Moreover, we explore the effect of the RNN's hidden state dimension (Dim) which affects the learning ability of the recurrent neural network.
As depicted in TABLE \ref{tab:d_comp}, the dimension of GRU cells is modified from 512 to 5,120.
The best performance is achieved at $Dim=4096$, which can better capture the spatial correlation of sequential views.
\begin{table}[htp]
\centering
\caption{The effect of the dimension (Dim) of GRU cells on the performance.}
\begin{tabular}{cccccc} \hline
Dim&512&1024&2048&4096&5120\\ \hline
Acc (\%) & 93.03 & 93.44 & 93.17 &  \textbf{93.99} & 92.76\\
\hline\end{tabular}
\label{tab:d_comp}
\end{table}
\begin{table}[htp]
    \centering
    \caption{The effect of the attention mechanism on the performance under airplane in FG3D-Net.}
    \begin{tabular}{ccccc} \hline
    Metric &OVA &OPA &NA &NR\\ \hline
    Acc (\%) & 93.31 & 93.58 & 92.90 &91.94 \\
    \hline\end{tabular}
    \label{tab:Att_comp}
\end{table}

\begin{table*}[htp]
    \centering
    \caption{We compare fine-grained shape classification accuracy(\%) for Airplane, Chair and Car categories in our FG3D dataset.}
    \label{table:compare}
    \begin{tabular}{l|c|c|cc|cc|cc}
    \hline
    \multirow{2}{*}{Method}&
        \multicolumn{1}{c|}{\multirow{2}{*}{Modality}} &
        \multicolumn{1}{c|}{\multirow{2}{*}{Input}} &
        \multicolumn{2}{c|}{Airplane} &
        \multicolumn{2}{c|}{Chair} &
        \multicolumn{2}{c}{Car}\\
        & \multicolumn{1}{c|}{}
        & \multicolumn{1}{c|}{}
        & \multicolumn{1}{c}{Class}
        & \multicolumn{1}{c|}{Instance}
        & \multicolumn{1}{c}{Class}
        & \multicolumn{1}{c|}{Instance}
        & \multicolumn{1}{c}{Class}
        & \multicolumn{1}{c}{Instance} \\ \hline
     PointNet\cite{qi2017pointnet}   &point &$1024 \times 3$ &82.67 &89.34 &72.07 &75.44 &68.12 &73.00 \\
     PointNet++ \cite{qi2017pointnet++} &point &$1024 \times 3$ &87.26 &92.21 &78.11 &80.78 &70.30 &75.21 \\
     SO-Net  \cite{li2018so}     &point &$1024 \times 3$ &66.10 &82.92 &62.78 &70.05 &53.38 &59.32 \\
     Point2Sequence \cite{liu2019point2sequence} &point &$1024 \times 3$ &87.52 &92.76 &74.90 &79.12 &64.67 &73.54 \\ 
     DGCNN \cite{wang2019dynamic} &point & $1024 \times 3$ &88.38 &93.17 &71.74 &77.51 &65.27 &73.61 \\ 
     RS-CNN \cite{liu2019relation} &point &$1024 \times 3$ &82.81 &92.35 &75.14 &78.96 &71.17 &77.11 \\ \hline
     VoxNet \cite{maturana2015voxnet} &voxel &$32^3$ & 84.45 &89.62 &66.50 &72.18 &63.75 &67.83 \\ \hline
     MVCNN \cite{Su_2015_ICCV}     &view  &$12 \times 224^2$ &82.57   &91.11 &76.27 &82.90 &71.88 &76.12 \\
     RotationNet \cite{kanezaki2018rotationnet} &view &$12 \times 224^2$ & 89.11 &92.76 &78.45 &82.07 &72.53 &75.59 \\
     3D2SeqViews \cite{han20193d2seqviews} &view &$12 \times 224^2$ &89.41 &92.21 &76.26 &77.10 &63.85 &66.39 \\
     3DViewGraph \cite{ijcai2019-0107} &view &$12 \times 224^2$ &88.21 &93.85 &79.88 &83.58 &71.65 &77.11 \\
     View-GCN \cite{wei2020view} &view &$12 \times 224^2$ &87.44 &93.58 &79.70 &83.63 &73.68 &77.34 \\
     SeqViews2SeqLabels \cite{han2018seqviews2seqlabels} &view &$12 \times 224^2$ &88.52 &93.44 &79.89 &82.54 &72.23 &75.36 \\
     Part4Feature \cite{han2019parts4feature} &view &$12 \times 224^2$ &82.55 &91.39 &77.08 &81.61 &73.42 &75.44 \\ \hline
     Ours       &view  &$12 \times 224^2$ &\textbf{89.44} &\textbf{93.99} &\textbf{80.04} &\textbf{83.94} &\textbf{74.03} &\textbf{79.47}\\
    \hline\end{tabular}
    \end{table*}
\begin{figure}
    \centering
    \includegraphics[width=8cm]{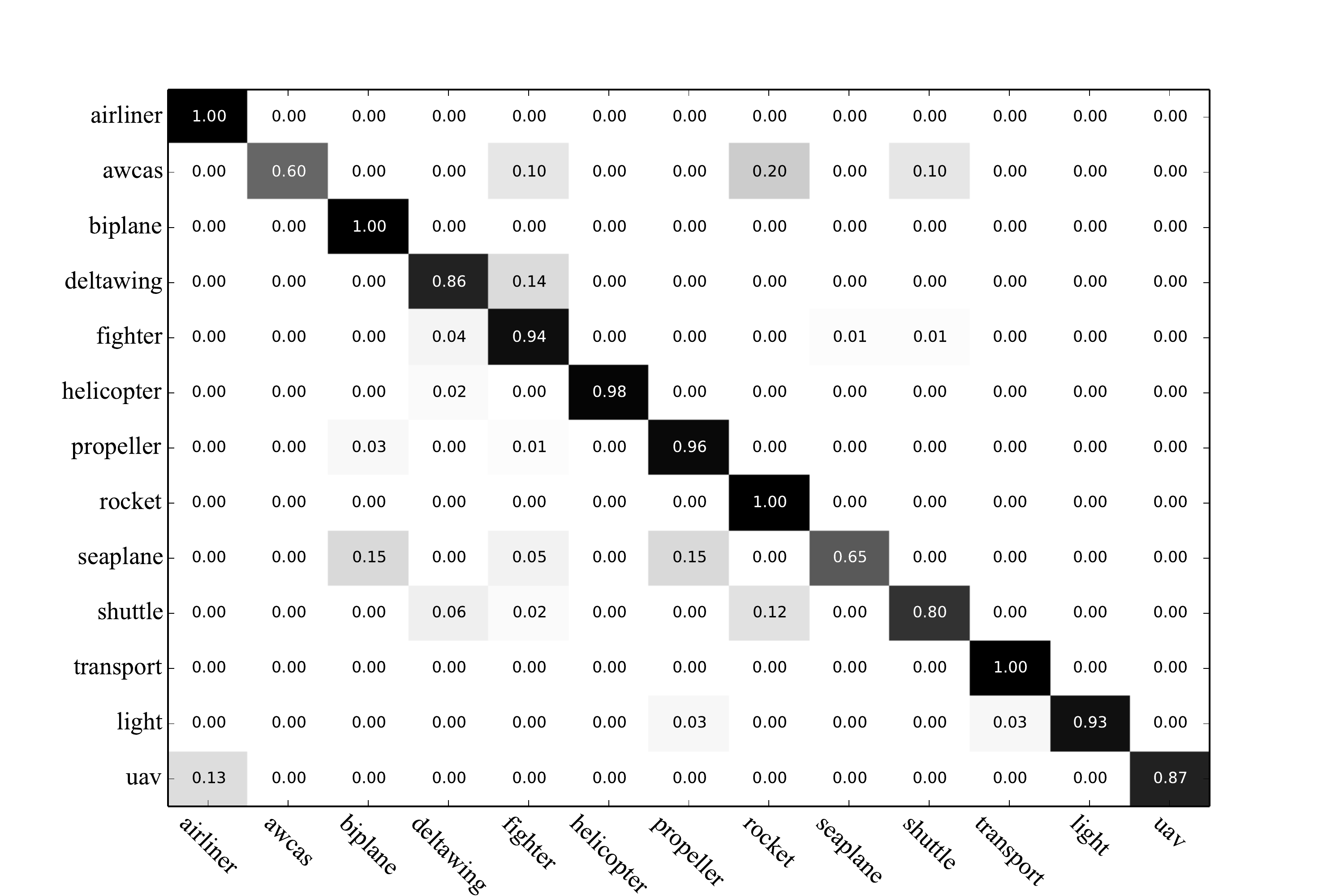}
    \caption{The classification confusion matrix under Airplane.}
    \label{fig:confusion_chair}
\end{figure}

\subsection{Ablation study}
In order to reveal the effect of the novel elements in FG3D-Net, such as attention mechanisms at different semantic levels, we performed ablation studies to justify their effectiveness.
We evaluate the performance of FG3D-Net with only part-level attention (OPA), only view-level attention (OVA), no attention (NA) or no RNN (NR).
Specifically, when we remove the attention mechanism, we set all attention values to the same constant.
For example, we set the part-level attention value to $\frac{1}{20}$ in OVA and the view-level attention value $\frac{1}{12}$ in OPA.
TABLE \ref{tab:Att_comp} illustrates the effectiveness of our attention mechanisms in learning highly discriminative representations for 3D shapes.
The results show that both part attention and view attention play an important role in extracting fine-grained details from multiple views.
Without the RNN layer, the performance of FG3D-Net drops significantly. 
This is mainly caused by the reduction of network parameters and the lack of spatial correlations among views.
In other words, the spatial correlation among views is important for the fine-grained classification of 3D shapes.

As for the components of neural networks, there is a Region Proposal Network in the GSP detection procedure, which contains a VGG-19 network, a RoI pooling layer and four FC layers. And for the subsequent networks, we select 20 GSPs from each view of total 12 views.
The GSPs detected from multiple views play an important role in capturing the fine-grained details of 3D shapes, but it also increases the complexity of our method. 
To better demonstrate the effectiveness of our method, we have evaluated the computational complexity of FG3D-Net on a NVIDIA GTX 1,080Ti GPU. 
Specifically, the model size and the average forward time are 832Mb and 358.19ms, respectively, where our FG3D-Net adopts 12 views of a 3D shape as input.

\subsection{Fine-grained visual classification}
We carry out the experiments in the fine-grained classification of 3D shapes under the proposed FG3D dataset with three categories including \textit{Airplane}, \textit{Car} and \textit{Chair}.
As shown in TABLE \ref{tab:fg3d_sta}, there are 4,173, 8,325 and 13,054 shapes in the categories, which are split into training and testing sets.
To evaluate the performance of FG3D-Net, we compare our method with several state-of-the-art 3D shape classifcation methods, which are trained under different 3D shape representations including point clouds, rendered views, and 3D voxels.
In TABLE \ref{table:compare}, we conduct the numerical comparisons including PointNet \cite{qi2017pointnet}, PointNet++ \cite{qi2017pointnet++}, Point2Sequence \cite{liu2019point2sequence}, DGCNN \cite{wang2019dynamic}, RS-CNN \cite{liu2019relation}, MVCNN \cite{Su_2015_ICCV}, SeqViews2SeqLabels \cite{han2019view}, RotationNet \cite{kanezaki2018rotationnet}, View-GCN \cite{wei2020view} and Parts4Feature \cite{han2019parts4feature}.
For these methods, we reproduce network structures using the public source code and evaluate them under our FG3D dataset.
\begin{figure}
    \centering
    \includegraphics[width=8cm]{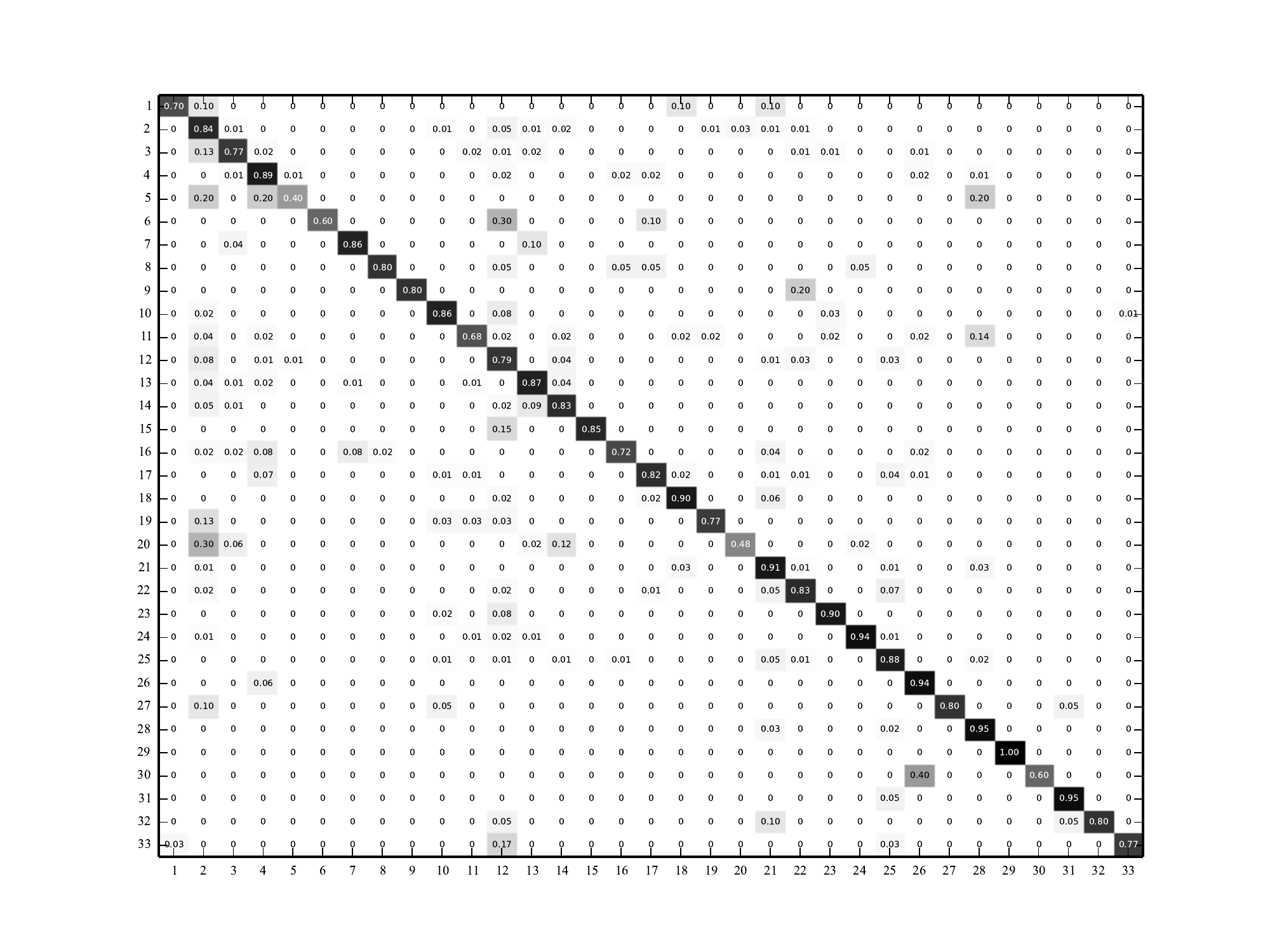}
    \caption{The classification confusion matrix under Chair.}
    \label{fig:confusion_airplane}
\end{figure}

In the comparison with other methods, we adopt MVCNN \cite{Su_2015_ICCV} as our baseline.
MVCNN is the pioneering study in 3D shape recognition from multiple views, which has achieved satisfactory performance in shape classifcation.
MVCNN leverages 12 views rendered around each 3D shape as input.
We follow this camera setting and evaluate the performance of other view-based methods, including RotationNet \cite{kanezaki2018rotationnet}, 3D2SeqViews \cite{han20193d2seqviews}, 3DViewGraph \cite{ijcai2019-0107}, View-GCN \cite{wei2020view}, SeqViews2SeqLabels \cite{han2018seqviews2seqlabels} and Parts4Feature \cite{han2019parts4feature}.
RotationNet learns the best camera setting to capture the view-specific feature representation for 3D shapes, which has obtained superior performance to previous state-of-the-art methods under ModelNet40 and ModelNet10.
Recent View-GCN \cite{wei2020view} also achieves high performances via view-based graph convolution network under ModelNet benchmarks. 
Therefore, we have compared FG3D-Net with RotationNet and View-GCN under the FG3D dataset using the same view sequence obtained from each 3D shape in this work.
To utilize the sequential views, SeqViews2SeqLabels adopts a recurrent neural network to aggregate multiple views of 3D shapes.
However, SeqViews2SeqLabels cannot explore the fine-grained information among GSPs extracted from the views.
Parts4Feature also detects generally semantic parts inside views similar to FG3D-Net.
However, Parts4Feature cannot fully explore the fine-grained correlation among features at different semantic levels. In particular, there is no view-level feature in Part4Feature, which results in a lack of view-level information in feature aggregation.

\begin{figure}
    \centering
    \includegraphics[width=8cm]{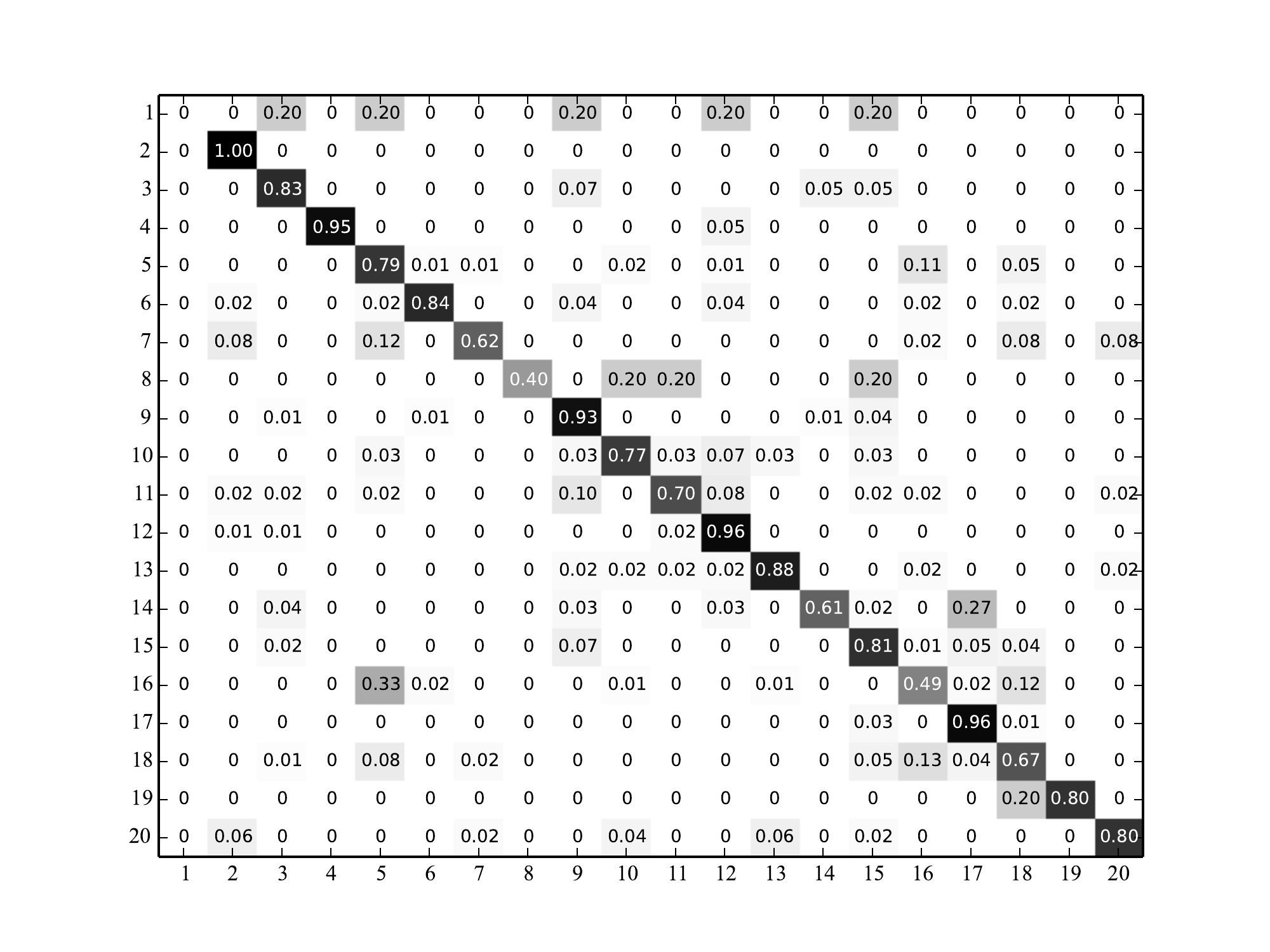}
    \caption{The classification confusion matrix under Car.}
    \label{fig:confusion_car}
\end{figure}
\begin{figure}
    \centering
    \includegraphics[width=8cm]{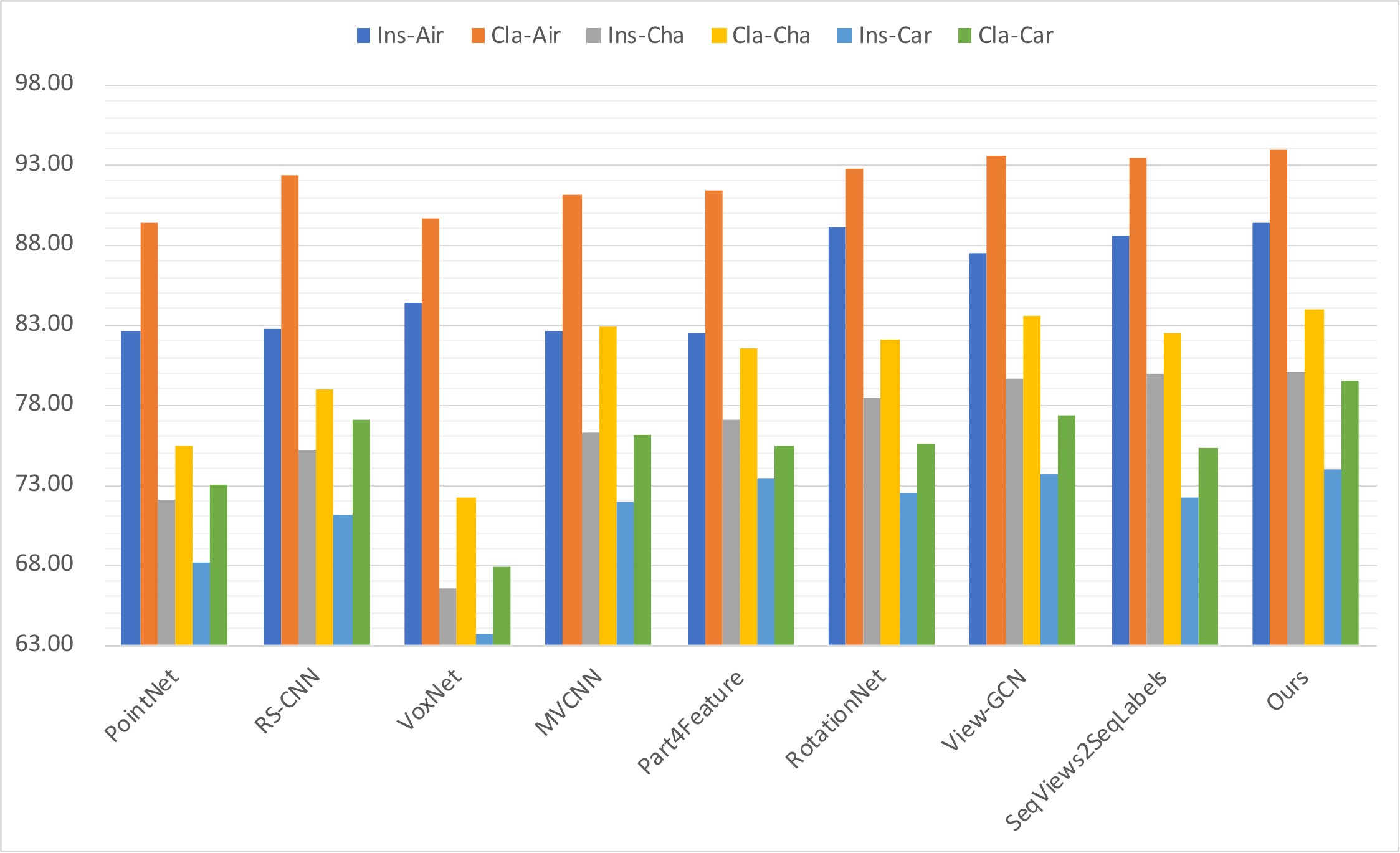}
    \caption{The intuitive display of classification accuracies under the FG3D Dataset, including instance accuracies (Ins-Air, Ins-Cha, Ins-Car) and class accuracies (Cla-Air, Cla-Cha, Cla-Car).}
    \label{fig:error_bar}
\end{figure}
\begin{figure*}
    \centering
    \includegraphics[width=14cm]{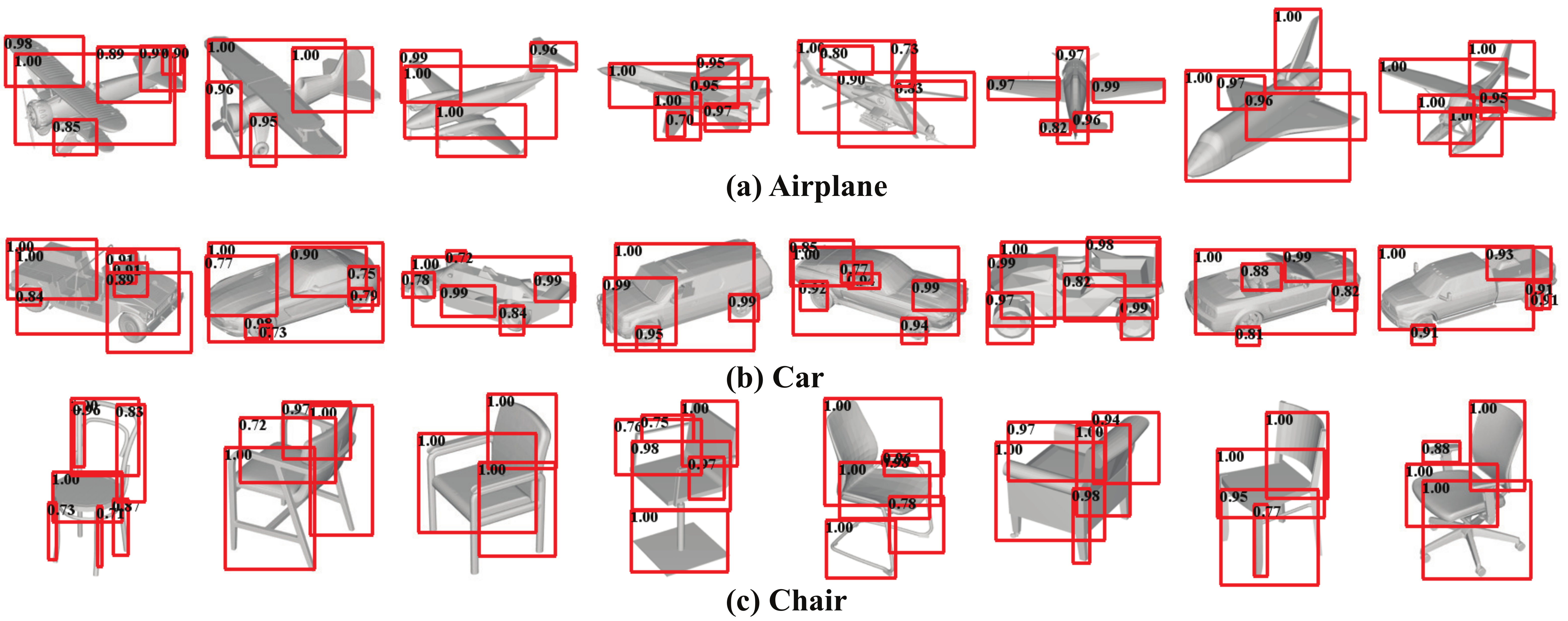}
    \caption{The results of generally semantic part detection, where the semantic score of each part is larger than 0.8.}
    \label{fig:detection}
\end{figure*}

\begin{figure*}
    \centering
    \includegraphics[width=14cm]{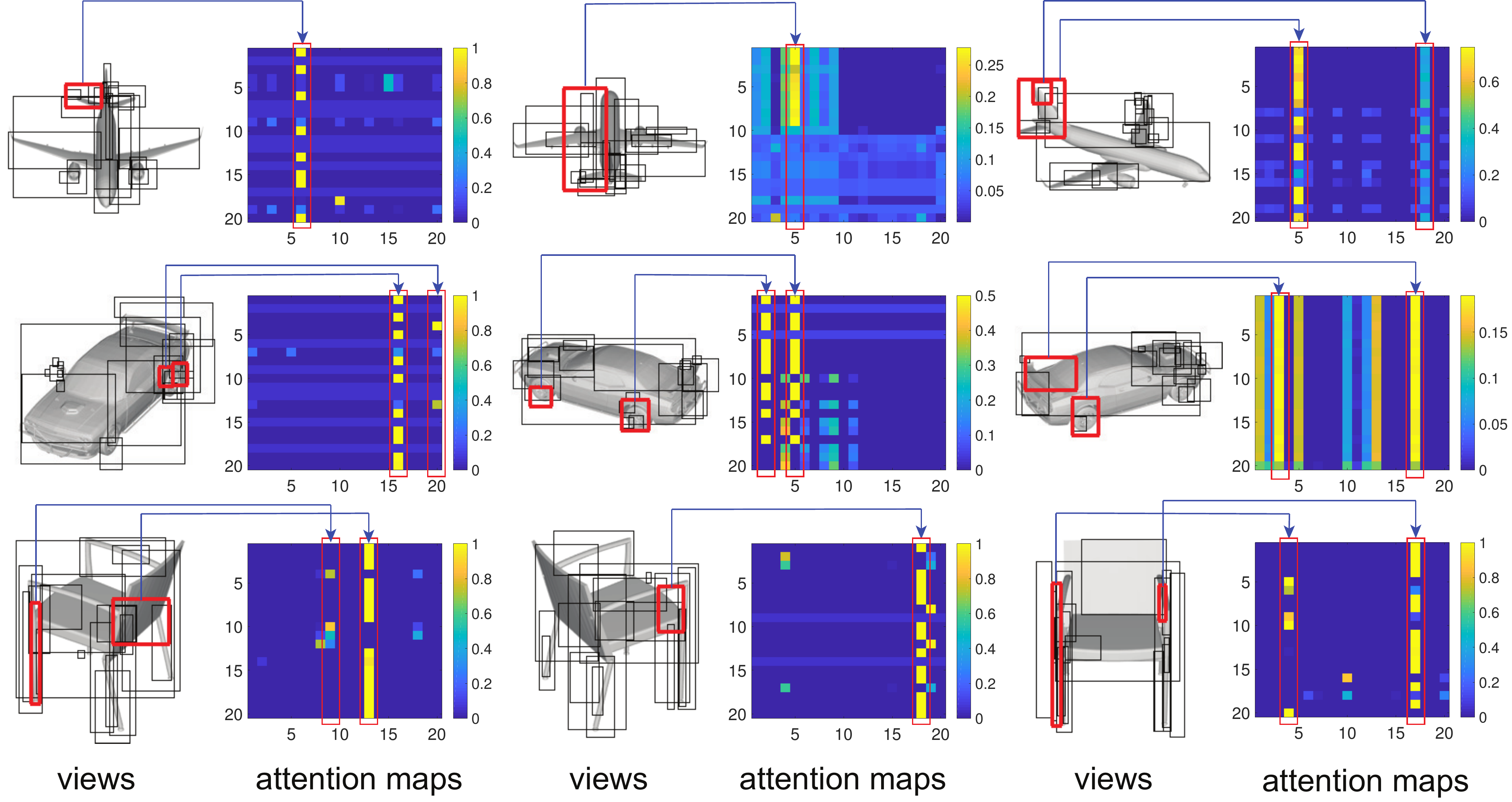}
    \caption{Visulization of the part-level attention mechanism. 
    There are 9 views from the 3 categories of our FG3D dataset, where the top 20 GSPs are drawn in each image. 
    The corresponding bilinear similarity attention map $q(k,k^{'})$ among GSPs in each view with a size of $20 \times 20$ is shown on the right of the view.
    We hightlight the discriminative GSPs with red bounding boxes and use arrows to indicate the corresponding columns on the attention maps.}
    \label{fig:vis_att}
\end{figure*}
In addition, we include methods that take other data formats as input.
For point-based methods, we choose the pioneering PointNet \cite{qi2017pointnet} and PointNet++ \cite{qi2017pointnet++} as the comparison targets.
We further include state-of-the-art methods such as SO-Net \cite{li2018so} and Point2Sequence \cite{liu2019point2sequence}.
To translate our FG3D dataset into point clouds, we apply Poisson Disk Sampling \cite{trimesh2017pointcloud}  to obtain 1,024 points for each 3D shape.
All point-based methods are trained with 1,024 points as input under FG3D.
VoxNet \cite{maturana2015voxnet}, also listed in TABLE \ref{table:compare}, is the poineering work in voxel-based methods for 3D shape recognition.

As shown in TABLE \ref{table:compare}, our FG3D-Net outperforms other state-of-the-art methods and achieves the highest classification accuracies in all three categories.
The results suggest that FG3D-Net can take advantage of generally semantic part detection to integrate fine-grained details in multiple views.
With a hierarchical aggregation strategy, we fully explore the correlation of features at different semantic levels.

To better demonstrate our classification results, we visualize the confusion matrix of our classification result under Airplane, Chair and Car in Fig. \ref{fig:confusion_airplane}, Fig. \ref{fig:confusion_chair} and Fig. \ref{fig:confusion_car}, respectively.
In each confusion matrix, an element in the diagonal line means the classification accuracy in a class, while other elements in the same row means the misclassification accuracy. The large diagonal elements show that FG3D-Net is good at classifying large-scale 3D shapes.
And to show the classification accuracies more intuitively, we draw the accuracy bar of FG3D-Net under different basic categories as illustrated in Fig. \ref{fig:error_bar}.
From the accuracy bar, FG3D-Net achieves higher performances than other compared methods.

\begin{figure}
    \centering
    \includegraphics[height=8cm]{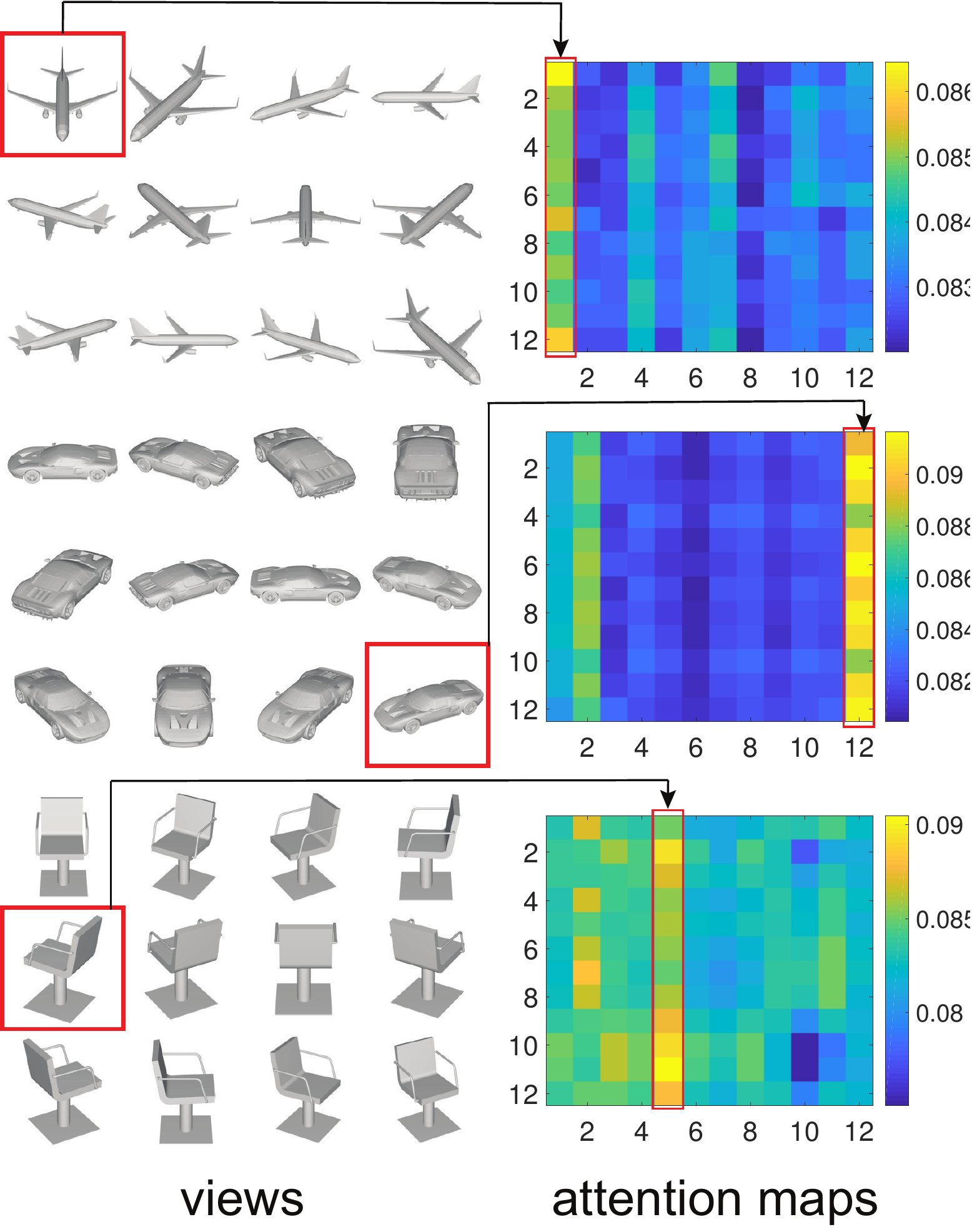}
    \caption{Visualization of the view-level attention mechanism. There are three 3D shapes selected from three different categories in our FG3D dataset. 
    In each row, we show the 12 views of each 3D shape on the left and the attention map $\theta(i,i^{'})$ of the views with a size of $12 \times 12$ on the right. 
    We hightlight the discriminative views with red bounding boxes and use arrows to indicate the corresponding columns on the attention maps.}
    \label{fig:vis_vtt}
\end{figure}

\section{Visualization}
In this section, we visualize some important properties of our FG3D-Net.
Firstly, we show examples of detected GSPs under our FG3D testing set.
In Fig. \ref{fig:detection}, we draw the bounding box of GSPs whose semantic score is larger than 0.8. 
We observe that FG3D-Net can extract the bounding boxes of discriminative GSPs, which supports representation learning of 3D shapes.

There are two attention mechanisms in FG3D-Net, including part-level and view-level attention.
To learn the global representation of 3D shapes, these attention mechanisms are important to preserve the fine-grained details inside GSPs from multiple views.
To intuitively show the effectiveness of these attention mechanisms, we draw some samples of part-level attention in Fig. \ref{fig:vis_att} and view-level attention in Fig. \ref{fig:vis_vtt}, respectively.
In Fig. \ref{fig:vis_att}, we show the attention map $q(k,k^{'})$ of 9 views from 3 different categories.
In each view, we draw the bounding boxes of top 20 GSPs, where the most discriminative GSPs with large attention values are in red.
We use arrows to indicate the correspondence of GSPs and attention values in the attention map $q(k,k^{'})$, where each column represents the attention value of each GSP.
Similar to the part-level attention, we visualize the view-level attention in Fig. \ref{fig:vis_vtt}, which shows the 12 views and the corresponding view attention map $\theta(i,i^{'})$ of a 3D shape in each row.
We also use arrows to indicate the correspondence of views and attention values on the attention maps.
The visualization results show that both part-level attention and view-level attention are effective to capture the fine-grained information in the feature aggregation.

\section{conclusions and future work}
We proposed FG3D-Net, a novel model to learn 3D gobal features via hierarchical feature aggregation from GSPs.
To evaluate the performance of FG3D-Net, we introduced a new fine-grained 3D shape classification dataset.
In the existing methods, the fine-grained details of generally semantic parts and the correlation of features in different semantic levels are usually ignored, which limits the discriminability of learned 3D global features.
To resolve these disadvantages, FG3D-Net employs a region proposal neural network to detect GSPs from multiple views, which identifies discriminative local parts inside views to learn the fine-grained deails of 3D shapes. 
In addition, we leverage part-level and  view-level attention mechanisms to effectively aggregate features in different semantic levels, which utilize the correlations among features to hightlight the discriminative features.
Finally, a recurrent neural network is adopted to capture the spatial information among views from multiple viewpoints, which takes advantage of prior information about the sequential input views.
Experimental results show that our method outperforms the state-of-the-art under the proposed 3D fine-grained dataset.

Although FG3D-Net learns 3D shape global features from GSPs to achieve high performance in fine-grained shape classification, it still suffers from three limitations.
First, FG3D-Net can only detect the limited types of GSPs, since the number of 3D shapes in the existing segmentation benchmarks is still limited. 
For example, the performance of our FG3D-Net may be further improved by integrating the fine-grained parts from PartNet \cite{mo2019partnet}.
Second, the number of categories in our FG3D dataset is somewhat small, where only three categories are currently included.
Thus, FG3D-Net merely  performs well under the current input setting and on a small number of shape categories, even with the help of bilinear similarity attention and RNN.
Third, there is no validation set in the spliting of our FG3D dataset.
A good validation set can help to obtain better hyperparameters in the network.
In the future, it is worth to explore unsupervised methods to detect GSPs inside multiple views and to further extend our FG3D dataset, including adding more categories and making a partition of validation set.
\bibliographystyle{IEEEtran}
\bibliography{IEEEabrv,reference}
\newpage
\vskip 0pt plus -1fil
\begin{IEEEbiography}[{\includegraphics[width=1in,height=1.25in,clip,keepaspectratio]{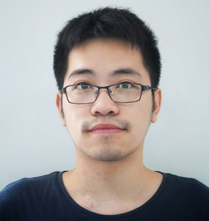}}]{Xinhai Liu}
    received the B.S. degree in computer science and technology from the Huazhong University of Science and Technology, China, in 2017. He is currently the PhD student with the School of Software, Tsinghua University. His research interests include deep learning, 3D shape analysis and 3D pattern recognition.
\end{IEEEbiography}
\begin{IEEEbiography}[{\includegraphics[width=1in,height=1.25in,clip,keepaspectratio]{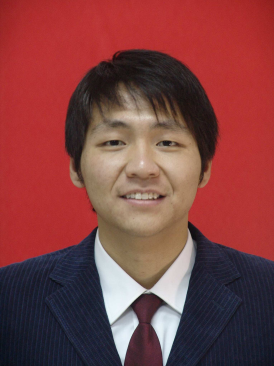}}]{Zhizhong Han}
    received the Ph.D. degree from Northwestern Polytechnical University, China, 2017. He is currently a Post-Doctoral Researcher with the Department of Computer Science, University of Maryland at College Park, College Park, USA. He is also a Research Member of the BIM Group, Tsinghua University, China. His research interests include machine learning, pattern recognition, feature learning, and digital geometry processing.
\end{IEEEbiography}
\begin{IEEEbiography}[{\includegraphics[width=1in,height=1.25in,clip,keepaspectratio]{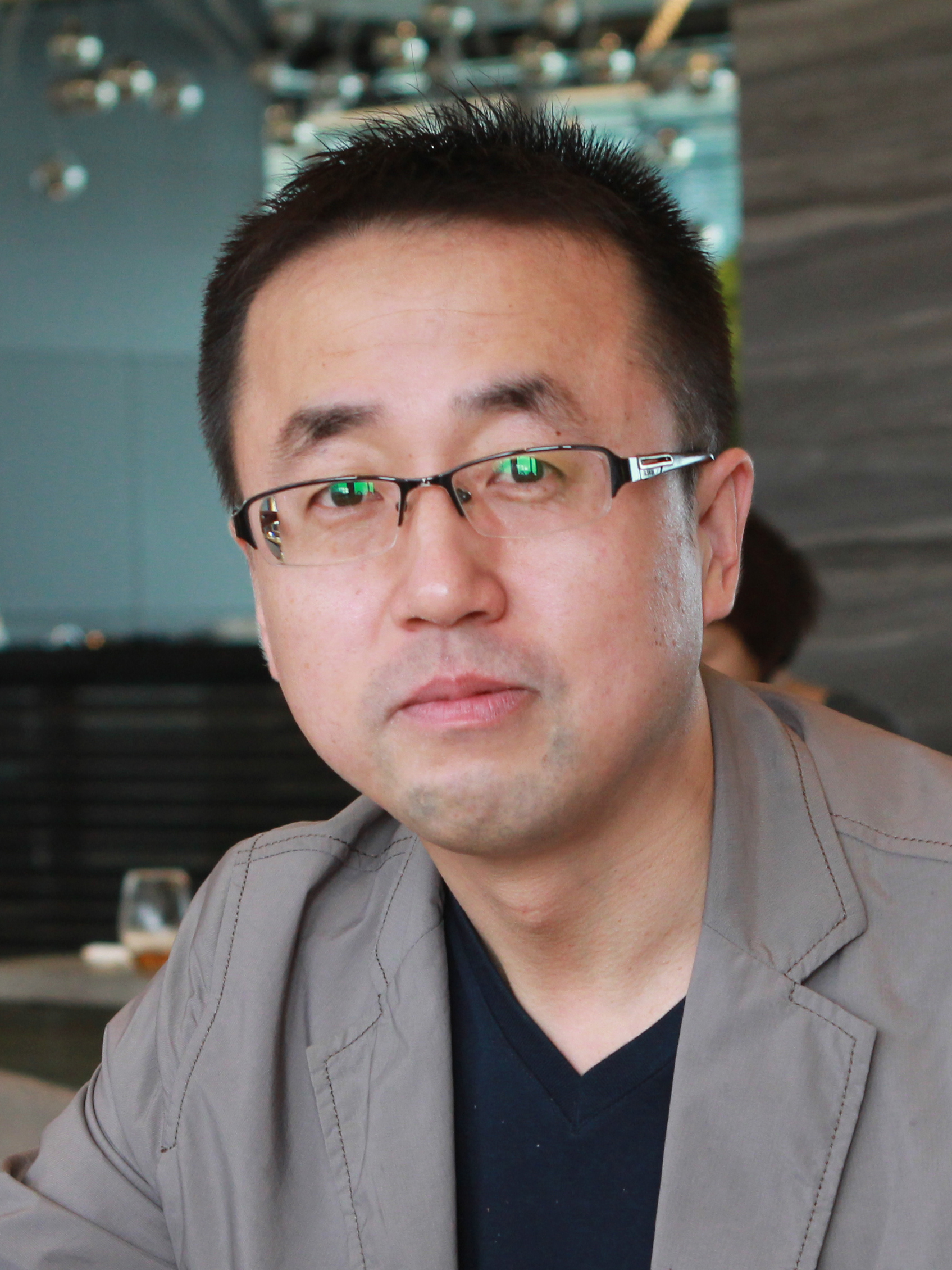}}]{Yu-Shen Liu}
    (M'18) received the B.S. degree in mathematics from Jilin University, China, in 2000, and the Ph.D. degree from the Department of Computer Science and Technology, Tsinghua University, Beijing, China, in 2006. From 2006 to 2009, he was a Post-Doctoral Researcher with Purdue University. He is currently an Associate Professor with the School of Software, Tsinghua University. His research interests include shape analysis, pattern recognition, machine learning, and semantic search.
\end{IEEEbiography}
\begin{IEEEbiography}[{\includegraphics[width=1in,height=1.25in,clip,keepaspectratio]{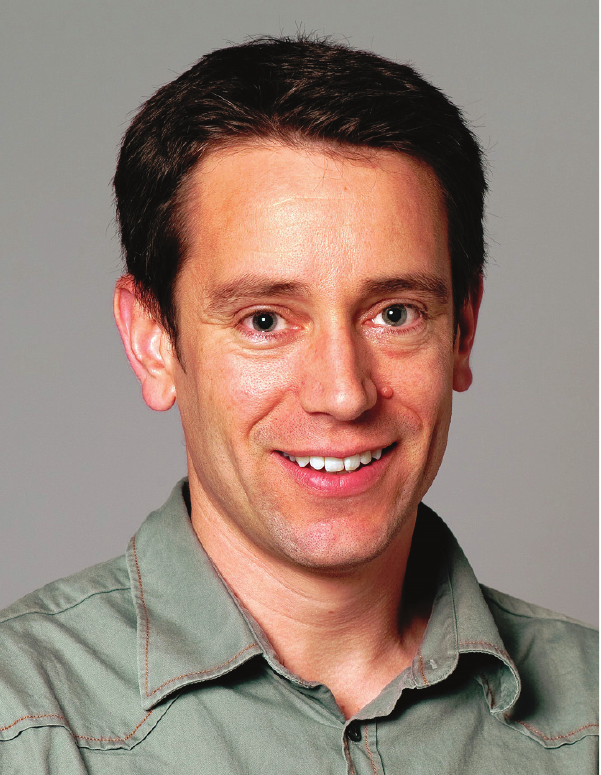}}]{Matthias Zwicker}
 is a professor at the Department
of Computer Science, University of Maryland, College Park, where he holds the Reginald Allan Hahne
Endowed E-nnovate chair. He obtained his PhD from
ETH in Zurich, Switzerland, in 2003. Before joining
University of Maryland, he was an Assistant Professor at the University of California, San Diego, and
a professor at the University of Bern, Switzerland.
His research in computer graphics covers signal
processing for high-quality rendering, point-based
methods for rendering and modeling, 3D geometry
processing, and data-driven modeling and animation.
\end{IEEEbiography}
\end{document}